\documentclass[pmlr,twocolumn,10pt]{jmlr}
\usepackage[T1]{fontenc}

\mlhtrack{proceedings}

\newif\iffinal
\finaltrue  % Use \finaltrue for camera-ready after acceptance

\iffinal
    \ifmlhneedspmlr
      \jmlrvolume{346}
      \jmlryear{2026}
    \fi
    \ifmlhfindings \jmlrproceedings{}{ML4H 2026 - Findings Track}\fi
    \ifmlhdemo     \jmlrproceedings{}{ML4H 2026 - Demo Track}\fi
    \jmlrworkshop{Machine Learning for Health (ML4H) 2026}
\else
    \jmlrproceedings{}{Submitted to ML4H 2026: \mlhtrackname}
    \jmlrworkshop{Machine Learning for Health (ML4H) 2026}
    
\fi

\usepackage{booktabs}
\usepackage{siunitx}
\usepackage[switch]{lineno}
\usepackage{makecell}
\usepackage{threeparttable}
\usepackage{microtype}
\usepackage{tikz}
\usepackage{minted}
\usepackage{booktabs, makecell, multirow}
\usepackage{jmlrutils}
\usepackage{placeins}
\usepackage{stfloats}

\newcommand{\yes}{\ensuremath{\bullet}}
\newcommand{\no}{\ensuremath{\circ}}
\newcommand{\rot}[1]{\rotatebox{60}{#1}}

\pgfdeclareplotmark{sqdiamond*}{%
  \pgfpathmoveto{\pgfqpoint{0pt}{-1.25\pgfplotmarksize}}%
  \pgfpathlineto{\pgfqpoint{1.25\pgfplotmarksize}{0pt}}%
  \pgfpathlineto{\pgfqpoint{0pt}{1.25\pgfplotmarksize}}%
  \pgfpathlineto{\pgfqpoint{-1.25\pgfplotmarksize}{0pt}}%
  \pgfpathclose
  \pgfusepathqfillstroke
}

\newcommand{\mrk}[2]{%
  \tikz[baseline=-0.6ex]{%
    \useasboundingbox (-#2,-#2) rectangle (#2,#2);
    \pgfsetplotmarksize{0.55ex}%
    \pgfuseplotmark{#1}%
  }%
}
\DeclareRobustCommand{\markercircle}{\mrk{*}{0.55ex}}
\DeclareRobustCommand{\markerdiamond}{\mrk{sqdiamond*}{0.55ex}}

\DeclareSIUnit{\million}{M}

\newcommand{\equal}[1]{{\hypersetup{linkcolor=black}\thanks{#1}}}

\title[UltraBench 2]{UltraBench 2: Towards Robust Evaluation of Vision Foundation Models on Ultrasound}

\author{
    \Name{Ashwath Radhachandran\nametag{\equal{These authors contributed equally.}}} \Email{ashwathradha123@g.ucla.edu} \\
    \addr Bioengineering Department, University of California, Los Angeles, CA, USA
    \AND
    \Name{Adam Tupper\nametag{\footnotemark[1]}} \Email{adam.tupper.1@ulaval.ca} \\
    \addr Institut Intelligence et Donn\'ees (IID), Universit\'e Laval, QC, Canada\\Mila -- Quebec AI Institute
    \AND
    \Name{Christian Gagn\'e} \Email{christian.gagne@gel.ulaval.ca} \\
    \addr Institut Intelligence et Donn\'ees (IID), Universit\'e Laval, QC, Canada\\Canada-CIFAR AI Chair, Mila -- Quebec AI Institute
    \AND
    \Name{William Speier} \Email{speier@ucla.edu} \\
    \addr Radiological Sciences Department, UCLA David Geffen School of Medicine \\ Bioengineering Department, University of California, Los Angeles, CA, USA
}

\begin{document}

\maketitle

\ifmlhdemo\else

\begin{abstract}
Benchmarking is an increasingly critical part of research in machine learning and the domains where it is applied, including healthcare. Yet, despite the steady development of new ultrasound foundation models in recent years, the development of well-designed benchmarks to evaluate them has lagged behind. This deficiency has led to fragmented and inconsistent evaluations of competing models, making it difficult to measure progress. To address this issue, we introduce UltraBench 2, a comprehensive benchmark with wide anatomical and task coverage, and a focus on standardization, reproducibility, and ease-of-use. Using this benchmark, we compare existing vision foundation models for ultrasound image analysis. Our analyses demonstrate that ultrasound-specific pretraining still leads on classification, but that state-of-the-art general-purpose models have drawn level on segmentation.
\end{abstract}
\fi

\ifmlhneedsstatements
\paragraph*{Data and Code Availability}

Datasets, models, code, and a public leaderboard are available at \href{https://github.com/adamtupper/ultrabench}{https://github.com/adamtupper/ultrabench}.

\paragraph*{Institutional Review Board (IRB)}
This work does not require IRB approval, as it uses only pre-existing, publicly available, de-identified datasets.
\fi

\section{Introduction}

Benchmarks are crucial for capturing the attention of researchers and driving progress on important problems \citep{hendrycks2024}. However, despite the steady development of new ultrasound foundation models, the establishment of well-designed benchmarks to evaluate them has lagged behind. Since the core value proposition of these models is that they are ``general-purpose'' and can be adapted to a wide variety of tasks with limited supervision, we require diverse benchmarks to evaluate their claims.

Well-designed benchmarks have several important characteristics. They should have clear and reproducible evaluation protocols to ensure fair comparisons, easy-to-use interfaces to minimize the barriers to entry, and a wide coverage of tasks to assess model flexibility. As summarized in Table 1 and discussed in Section \ref{sec:related-work-benchmarks}, existing medical image analysis benchmarks satisfy some, but not all of these criteria. This limitation makes it difficult to reliably measure progress in ultrasound analysis.

In this paper, we propose UltraBench 2, a new benchmark for ultrasound image analysis. This benchmark builds upon the foundations of existing benchmarks \citep{jiang2025,ma2026,tupper2025} by expanding task coverage, improving usability, and standardizing dataset preprocessing and model evaluation.

Using UltraBench 2, we perform a retrospective analysis of existing ultrasound foundation models and identify areas in which they excel and struggle. We find that existing models tend to struggle more at pathological tasks (e.g., segmenting pathological structures and diagnosing malignancy) than non-pathological tasks (e.g., segmenting anatomical structures or identifying fetal anatomical planes). We also find that state-of-the-art general-purpose vision foundation models, specifically DINOv3 \citep{simeoni2025},  have closed the gap on image segmentation performance. These insights provide focus for the development of future models and highlight promising avenues for improvement.

The remainder of this paper is organized as follows. First, we compare UltraBench 2 to previous benchmarks and describe the current landscape of ultrasound foundation models. Then, in Section \ref{sec:ultrabench2}, we describe the design of the benchmark and its tasks. Finally, we perform a retrospective analysis of existing models using the benchmark, present our findings, and discuss the implications for future work on the development of ultrasound foundation models.

\section{Related Work}
\label{sec:related-work}

\begin{table*}[ht]
\centering
\caption{A comparison between ultrasound-specific and general biomedical image analysis benchmarks. UltraBench 2 is the only one that integrates dataset downloading, standardized preprocessing and evaluation, a diverse coverage of ultrasound tasks and anatomical regions, and permissive licensing.}
\label{tab:benchmarks}
\footnotesize
\resizebox{\textwidth}{!}{%
\begin{threeparttable}
\begin{tabular}{@{}lcccccccc@{}}
\toprule
\multirow{2}{*}{\thead{Benchmark}}
  & \multicolumn{2}{c}{\thead{Ultrasound tasks}}
  & \multirow{2}{*}{\thead{Anatomical\\regions\tnote{$\dagger$}}}
  & \multirow{2}{*}{\thead{Standardized\\preprocessing}}
  & \multirow{2}{*}{\thead{Train/test\\splits}}
  & \multirow{2}{*}{\thead{Download\\interface}}
  & \multirow{2}{*}{\thead{Licensing}}
  & \multirow{2}{*}{\thead{Leaderboard}} \\
\cmidrule(lr){2-3}
  & \thead{Class.} & \thead{Seg.} & & & & & & \\
\midrule
\textit{Biomedical image} \\
MedMNIST v2  & 1  & --  & 1  & Yes  & Yes  & Yes   & Non-commercial & No  \\
MedSegBench  & -- & 5   & 5  & Yes  & Yes  & Yes   & Non-commercial & No  \\
MedIMeta     & 2  & --  & 1  & Yes  & Yes  & Yes   & Non-commercial & No  \\
\addlinespace
\textit{Ultrasound-specific} \\
UniUS-Bench  & 8  & 10  & 12 & No   & No   & No    & Restricted     & No  \\
UltraBench   & 7  & 7   & 10 & Code & Code & No    & Restricted     & No  \\
UltraFedFM   & 10 & 5   & 13 & No   & No   & No    & Non-commercial & No  \\
US-43d       & -- & 43  & 13 & Code & No   & No    & Restricted     & No  \\
\midrule
UltraBench 2 & 11 & 10 & 10 & Yes  & Yes  & Yes & Non-commercial & Yes \\
\bottomrule
\end{tabular}
\begin{tablenotes}
\item[$\dagger$] See Appendix~\ref{app:anatomical-regions} for details on how anatomical regions are counted.
\end{tablenotes}
\end{threeparttable}%
}
\end{table*}

\subsection{Benchmarks in Medical Imaging}
\label{sec:related-work-benchmarks}

Good examples of multi-domain biomedical imaging benchmarks are MedMNIST v2 \citep{yang2023}, MedSegBench \citep{kus2024}, and MedIMeta \citep{woerner2025}. These benchmarks satisfy many of the important qualities described in the introduction: they standardize the format of the individual datasets and the preprocessing applied to images, define evaluation metrics, and provide user-friendly interfaces for downloading and working with the data. However, these datasets include few ultrasound tasks.

The few existing ultrasound-specific benchmarks \citep{jiang2025,ma2026,tupper2025} each have gaps, as shown in \tableref{tab:benchmarks}. Each one covers a good range of anatomical regions and tasks, but they fall short on reproducibility and usability.

No benchmark provides fixed train and test splits directly. UltraBench provides code to generate the splits, using fixed random seeds. But this still cannot guarantee the exact same split on a different machine or software version. The remaining benchmarks do not release the code for splitting at all, so the original splits cannot be recovered.

The same problem applies to preprocessing. UltraBench and US-43d release the code used to prepare the images. This adds work for other researchers, but the steps can still be reproduced. UniUS-Bench and UltraFedFM do not. For these benchmarks, the exact images cannot be reproduced at all, making reproducibility impossible (without help from the original authors) rather than just inconvenient.

Existing benchmarks also only link to the original datasets. Researchers must download each dataset separately and convert it to a usable format themselves. This creates further room for differences between setups. UniUS-Bench, UltraBench, and US-43d also include datasets with narrow, competition-specific licenses \citep{butterflynetwork2024, morelia2016}, which could limit their use. All other benchmarks use datasets with non-commercial or more open licenses.

Together, these problems make results hard to replicate and the benchmarks hard to use. We argue this has led to the current situation of models being tested on different sets of tasks and compared against different sets of competing models (see Appendix \ref{app:ultrasound-foundation-models}). U2-Bench \citep{le2026} was recently released to address similar issues for large vision-language models, but it is not designed for vision foundation models.

\subsection{Ultrasound Foundation Models}
\label{sec:foundation-models}

There has been a steady stream of ultrasound foundation model releases since 2023. These models have applied and adapted self-supervised learning techniques developed for general-purpose vision foundation models, often to train vision transformers (ViTs) \citep{dosovitskiy2021} with modifications to address ultrasound-specific image characteristics, such as low signal-to-noise ratio.

Nearly all models use transformer-based architectures. USFM \citep{jiao2024}, URFM \citep{kang2025}, USF-MAE \citep{megahed2026}, UltraFedFM \citep{jiang2025}, and TinyUSFM \citep{ma2026} all use variants of the ViT architecture. Both ViT and ConvMAE \citep{gao2022a} models were released for DeblurringMIM \citep{kang2024a}. EchoCare \citep{zhang2025} is a Swin Transformer model \citep{liu2021}. UltraSam \citep{meyer2025} and SAMUS \citep{lin2024c} use Segment Anything Model (SAM)-style ViTs \citep{kirillov2023}, but SAMUS uses a pretrained ViT backbone with a parallel CNN to adapt the model to ultrasound. Finally, OpenUS \citep{zheng2025} uses the VMamba architecture \citep{liu2024b}.

The majority of ultrasound foundation models (USFM, DeblurringMIM, URFM, USF-MAE, EchoCare, and UltraFedFM) are pretrained via masked image modeling \citep{xie2022a,he2022}. USFM introduces a dual spatial-frequency masking scheme to improve robustness to low-quality images, while DeblurringMIM blurs its inputs, challenging reconstruction to recover fine-grained diagnostic detail. URFM and TinyUSFM instead target learned representations rather than pixels. URFM reconstructs BiomedCLIP features \citep{zhang2025a}, integrating additional medical insight, and TinyUSFM distills a pretrained USFM into a ViT-Tiny student using a curated pretraining dataset and a distillation loss designed to prevent collapse. EchoCare adds a second, anatomy-classifier decoder so that representations encode hierarchical anatomical relationships. SAMUS and UltraSam are the only models to have been pretrained using supervised training for semantic segmentation.

Across these models, pretraining data is drawn from public sources, private repositories, or a mixture of both. Dataset sizes range from tens of thousands (SAMUS) to millions (UltraFedFM, USFM, and EchoCare) of images. Most models are trained on diverse multi-organ, multi-center, and multi-device sets of images. A notable exception is DeblurringMIM, which was trained exclusively on 280,000 thyroid ultrasound images.

In addition to ultrasound foundation models, we also benchmark several ViT and ConvNeXt \citep{liu2022e} DINOv3 models \citep{simeoni2025} as references for the performance of state-of-the-art general purpose image foundation models. Further details about each model and its pretraining data are included in Appendix \ref{app:ultrasound-foundation-models}.

\section{UltraBench 2}
\label{sec:ultrabench2}

\subsection{Design Principles}

The design of UltraBench 2 follows principles drawn from established guidance on benchmark construction \citep{hendrycks2024, maier-hein2020,reuel2024,ott2022} and the shortcomings of the existing benchmarks discussed in \sectionref{sec:related-work-benchmarks}.

\paragraph{Broad coverage.} Model rankings are sensitive to the choice of evaluation tasks \citep{dehghani2021,kerssies2024}, so a single dataset or task type is insufficient for model selection. UltraBench 2 spans both segmentation and classification across 10 anatomical regions, five clinical application areas, and multiple transducer types. Similar to \citet{simumba2026}, we group tasks into categories to report capability-level scores (classification vs. segmentation; pathological vs. anatomical targets) that show where models are strong or weak.

\paragraph{Standardization.} We publish fixed train/test splits, preprocessed test sets, and evaluation scripts to ensure that submissions are evaluated on identical inputs using identical evaluation protocols. Following the recommendations of \cite{jimenez-sanchez2024}, details of how each task/dataset is prepared are included in Appendix \ref{app:datasets}.

\paragraph{A low barrier to entry.} Every dataset is exposed through a unified interface built on Hugging Face Datasets \citep{lhoest_datasets_2021} to maximize the ease of use and interoperability with different deep learning frameworks. An example of this interface is shown in Listing \ref{listing:interface}. We also provide reference PyTorch implementations of existing models and task-specific heads. To ensure consistency and to streamline submissions, we provide framework-independent evaluation scripts that calculate the benchmark metrics based on CSV prediction files (i.e., per-image/sequence/patient labels for classification tasks and run-length encoded (RLE) strings for segmentation tasks), similar to challenges hosted on Kaggle and Grand Challenge.

\begin{listing}[t]
\begin{minted}[fontsize=\small]{python}
from ultrabench2 import load_dataset
from torch.utils.data import DataLoader

# A hosted dataset
dataset = load_dataset(name="AUL")

# Or, a local dataset
dataset = load_dataset(
    name="OpenKidney",
    path="path/to/raw/data",
)

# Use with any framework (e.g., PyTorch)
loader = DataLoader(dataset["test.256"])
\end{minted}
\caption{A unified interface for using hosted and local datasets, backed by Hugging Face Datasets.}
\label{listing:interface}
\end{listing}

\paragraph{Longevity.} Benchmarks lose their discriminative power as they saturate, and design choices at construction time play a part in how quickly this happens \citep{akhtar2026,ott2022}. We screened candidate datasets for reported performance ceilings to try to retain tasks with room for improvement. We also restrict the number of training examples for each task to a maximum of 1000 patients, sequences, or images (depending on the granularity of the labels). Longevity also depends on stewardship. We host mirrors of the source datasets where permitted under their licenses and maintain a public leaderboard restricted to reproducible open-source submissions.

\subsection{Tasks}
\label{sec:tasks}

\begin{table*}[t]
  \centering
  \caption{Evaluation tasks in UltraBench~2. Frame counts are given as train/test; the
  training split is capped at 1000 examples per task. For tasks with patient- or sequence-level labels, one example is one patient or sequence, respectively.
  Class counts include the background class for segmentation tasks.}
  \label{tab:tasks}
  \footnotesize
  \setlength{\tabcolsep}{3pt}
  \resizebox{\textwidth}{!}{%
  \begin{tabular}{@{}lclllcrl@{}}
    \toprule
    Task & Classes & Dataset & Anatomy & Transducer & Frames (train/test) & Patients & Labels \\
    \midrule
    \multicolumn{8}{@{}l}{\textit{Pathological classification}} \\
    \quad Breast tumor pathology & 2 & BUS-BRA \citep{gomez-flores_bus-bra_2024} & Breast & Linear & 1000/875 & 1064 & Frame \\
    \quad Breast tumor histology & 9 & BUS-BRA \citep{gomez-flores_bus-bra_2024} & Breast & Linear & 1000/875 & 1064 & Frame \\
    \quad COVID-19 & 2 & COVID-BLUES \citep{wiedemann_covid-blues_2025} & Lung & Convex & 21528/10218 & 63 & Patient \\
    \quad Gallbladder tumor & 3 & GBCU \citep{basu_surpassing_2022} & Gallbladder & Convex & 1000/122 & 218 & Frame \\
    \quad Liver mass & 3 & AUL \citep{xu_improving_2023} & Liver & Convex & 588/147 & 735 & Frame \\
    \quad MASLD & 2 & MASLD \citep{byra_transfer_2018} & Liver & Convex & 440/110 & 55 & Patient \\
    \quad Ovarian tumor & 8 & MMOTU-2D \citep{zhao_mmotu_2023} & Ovary & Convex & 1000/469 & 247 & Frame \\
    \quad Thyroid nodule & 2 & TN3K \citep{gong_thyroid_2023} & Thyroid & Linear & 1000/614 & 1119 & Frame \\
    \addlinespace
    \multicolumn{8}{@{}l}{\textit{Non-pathological classification}} \\
    \quad Cardiac image quality & 3 & CAMUS \citep{leclerc_deep_2019} & Heart & Phased array & 17264/1968 & 500 & Sequence \\
    \quad Fetal brain plane & 4 & Fetal Planes \citep{burgos-artizzu_evaluation_2020} & Fetus & Convex & 1000/2092 & 1082 & Frame \\
    \quad Fetal plane & 6 & Fetal Planes \citep{burgos-artizzu_evaluation_2020} & Fetus & Convex & 1000/11400 & 1792 & Frame \\
    \midrule
    \multicolumn{8}{@{}l}{\textit{Pathological segmentation}} \\
    \quad Breast tumor & 2 & BUS-BRA \citep{gomez-flores_bus-bra_2024} & Breast & Linear & 1000/875 & 1064 & Frame \\
    \quad Liver mass & 2 & AUL \citep{xu_improving_2023} & Liver & Convex & 588/147 & 735 & Frame \\
    \quad Ovarian tumor & 2 & MMOTU-2D \citep{zhao_mmotu_2023} & Ovary & Convex & 1000/469 & 247 & Frame \\
    \quad Thyroid nodule & 2 & TN3K \citep{gong_thyroid_2023} & Thyroid & Linear & 1000/614 & 1119 & Frame \\
    \addlinespace
    \multicolumn{8}{@{}l}{\textit{Anatomical segmentation}} \\
    \quad Liver & 2 & AUL \citep{xu_improving_2023} & Liver & Convex & 587/147 & 734 & Frame \\
    \quad Cardiac region & 4 & CAMUS \citep{leclerc_deep_2019} & Heart & Phased array & 1000/1968 & 500 & Frame \\
    \quad Kidney capsule & 2 & Open Kidney \citep{singla_open_2023} & Kidney & Convex & 411/103 & 514 & Frame \\
    \quad Kidney regions & 4 & Open Kidney \citep{singla_open_2023} & Kidney & Convex & 411/103 & 514 & Frame \\
    \quad Pubic symphysis--fetal head & 3 & PSFHS \citep{chen_psfhs_2024} & Pelvis, fetus & Convex & 1000/358 & 1124 & Frame \\
    \quad Muscle & 2 & STMUS \citep{marzola_deep_2021} & Muscle & Linear & 1000/7169 & 1223 & Frame \\
    \bottomrule
  \end{tabular}%
  }
\end{table*}

UltraBench 2 includes the 10 semantic segmentation and 11 image classification tasks listed in Table \ref{tab:tasks}. These tasks cover clinical and non-clinical applications: cancer screening, prenatal care, anatomical region segmentation, disease diagnosis, and image quality assessment. Detailed descriptions of each task are included in Appendix \ref{app:datasets-summary}.

\subsection{Dataset Curation and Processing}

The 12 datasets in UltraBench 2 are curated from existing publicly available sources with non-commercial or more open licenses. When considering datasets for inclusion, we took particular care to exclude those with previously raised concerns about data quality (e.g., BUSI \citep{al-dhabyani2020,pawlowska2023}), those containing heavily annotated images  (e.g., LEPset \citep{li2024e}), and those with potential licensing issues (e.g., Butterfly \citep{butterflynetwork2024}, POCUS \citep{born2021}, and US Nerve \citep{morelia2016}). We also preferred smaller datasets (e.g., TN3K \citep{gong_thyroid_2023} over Stanford Thyroid \citep{stanfordaimicenter2021}) where multiple suitable options were available, to avoid overly limiting pretraining corpora.

For the selected datasets, the images are divided into training and test splits using stratified sampling where applicable to ensure similar label distributions between the two splits. For datasets with multiple images per patient, the dataset is split by patient. Since the datasets contain a mixture of RGB, RGBA, and greyscale images, we convert all images to RGB by either stacking single channel images, or removing the alpha channel. Other than standardizing the number of channels, no further pre-processing (e.g., resizing) is applied to the training images. This provides maximum freedom for new submissions to innovate on both data augmentation and preprocessing as well as model architecture and training algorithms. For the test set, two versions are published. One version with original-sized images (with the same minimal preprocessing as the training set) and another copy with the images resized to $256 \times 256$ pixels using ``letterbox'' resizing to preserve their original aspect ratios. Performance is reported on the resized version to ensure that comparisons between models are fair.

We provide hosted versions of datasets that we have licenses to redistribute on the Hugging Face Hub. These datasets are either distributed in their preprocessed form, for those that allow derivatives, or preprocessed (and cached) by the package at runtime for datasets under non-derivative licenses. The Open Kidney and GBCU datasets are the only two that are gated behind research use agreements. For these datasets, users provide a path to the raw data, which is then processed and cached in the same manner as a hosted non-derivative dataset.

\subsection{Model Evaluation}

Performance on classification and segmentation tasks is measured using macro F1 and positive Dice scores, respectively. The positive Dice score excludes the background class since the object(s) of interest typically represent only a small fraction of the image and including the background would inflate results.

The individual task scores are aggregated into four category scores: anatomical segmentation, pathological segmentation, pathological classification, and non-pathological classification. This distinguishes a model's ability to segment anatomical structures (e.g., organs) and pathological structures (e.g., nodules), as anatomical structures tend to be larger and more clearly defined than pathological ones. It also separates diagnostic tasks from image quality and categorization tasks. For tasks with patient- or sequence-level labels, performance is measured per-patient or per-sequence.

Each submission must provide test set predictions for $S = 5$ different seeds per task. Let $Y_{k,r}$ be the performance metric of task $k$ under run $r$, and let a category consist of $K$ tasks scored over $S$ runs. The reported category score is the unweighted mean of the run-averaged task scores:

\begin{equation}
    \bar{Y} = \frac{1}{K}\sum_{k=1}^{K}\bar{Y}_k,\qquad \bar{Y}_k = \frac{1}{S}\sum_{r=1}^{S} Y_{k,r}.
\end{equation}

Uncertainty is quantified using error bars, $\bar{Y} \pm \mathrm{SE}(\bar{Y})$, which combine two sources of noise: the variability across runs and the sampling noise in the test sets. Both are computed at the level of the category score rather than per task, so that correlations between tasks that share a test set (e.g., BUS-BRA pathology and histology classification) are not ignored. $\mathrm{SE}(\bar{Y})$ is calculated as

\begin{equation}
    \mathrm{SE}(\bar{Y}) = \sqrt{\frac{\sigma_{\mathrm{run}}^{2}}{S} + \sigma_{\mathrm{boot}}^{2}},
\end{equation}

where $\sigma_{\mathrm{run}}$ is the standard deviation across runs of the per-run category score, and $\sigma_{\mathrm{boot}}$ is the standard deviation of the category score across 2000 bootstrap replications. Each replication resamples the test set of each dataset and recomputes the category score with each task's metric averaged over the five runs. Where a dataset provides per-image patient identifiers, resampling is done at the patient level, since frames from one patient are correlated.

Because we fix the train/test split, these standard errors are conditional on this split. They quantify sampling variability of the test cohorts and stochasticity of training, not variability that would arise from a different split, different hyperparameter search, or a different source population.

\subsection{Leaderboard}
\label{sec:leaderboard}

To track the progress of ultrasound foundation models, we provide a leaderboard for open-source models and encourage users to evaluate their models on different categories. We restrict the leaderboard to open-source models to mitigate the risk of evaluations being performed in bad faith, as the datasets are public and there are no hidden/private test sets that could be used to prevent bad actors from training on the test data.

\section{Evaluations of Existing Models}

We performed a retrospective analysis of existing ultrasound foundation models to populate the benchmark. It is important to note that some of the datasets used in the benchmark were used in the pretraining phase of existing models. The extent of this overlap is documented in Appendix \ref{app:pretraining-overlap}.

\subsection{Evaluation Setup}

Each model described in \sectionref{sec:foundation-models} was evaluated via full fine-tuning and head-only training (i.e., frozen backbone). Full fine-tuning has a higher performance potential as it allows models to adapt the learned features to the specific task, at the cost of higher computational requirements and the potential to overfit on small datasets. Head-only training is computationally cheaper and limits the risk of overfitting, but relies on the model being able to extract relevant features for the task without adaptation.

For both training regimes, the models were trained for a maximum of 10,000 steps using AdamW, a batch size of 32, and a cosine annealing learning rate scheduler with an initial learning rate of $\eta_\text{head} = 1\text{e}^{-3}$. The models were evaluated every 100 steps on the validation set, with early stopping if there was no reduction in the validation loss after 10 validation checks (i.e., 1,000 training steps). For full fine-tuning, the backbone was frozen for the first 1,000 training steps to allow the randomly-initialized head to warm up against fixed pretrained features. The backbone learning rate was then linearly warmed up from zero over the following 1,000 steps to a ceiling of $r_\text{max} = 0.01 \times \eta_\text{head}$. For full fine-tuning, we also applied layer-wise learning rate decay \citep{howard2018} to lower the learning rates for shallower layers and TrivialAugment-style data augmentation ~\citep{muller2021} that randomly applied one of identity, rotation (up to $30^{\circ}$), translation, random resized crop (scale $0.6$--$1.0$), Gaussian noise, or brightness/contrast jitter. The full training details and hyperparameters are provided in \appendixref{app:training-setup}.

\paragraph{Classification.} For classification tasks, we attached a single linear layer mapping the encoder's final $d$-dimensional feature embedding to $C$ class logits. For isotropic ViT backbones that produce $d$-dimensional patch token embeddings $[B, N, d]$ (TinyUSFM, USFM, URFM-B/L, UltraFedFM, DeblurringMIM, DINOv3-ViT-B/L, and UltraSAM), these were mean-pooled to produce the final embedding vector. For models with a $d$-dimensional \texttt{[CLS]} token (USF-MAE) we used this directly. For models that produce a spatial feature map (EchoCare, OpenUS, DINOv3 ConvNeXt, and SAMUS), this was mean-pooled and flattened into a $d$-dimensional embedding. Cross-entropy loss was used for all tasks. For patient- and sequence-level tasks, we evaluated each model using the average predicted probability across all images belonging to the patient or sequence, respectively. The parameter counts for each combination of model and head are reported in Table \ref{tab:cls_param_count} of the appendix.

\paragraph{Segmentation.} The segmentation decoder was selected based on each backbone's output structure. For models producing patch token sequences or spatial feature maps (all ViT-based and SAM-based models), we used a progressive upsampler (PUP) decoder~\citep{zheng_rethinking_2021}, which reshapes patch tokens $[B, N, d]$ onto a 2D grid and recovers the target resolution through a sequence of transposed-convolution stages. This decoder was also used for SAMUS and UltraSam because their SAM-based architectures do not expose intermediate transformer blocks. For hierarchical multi-scale backbones (EchoCare, OpenUS, DINOv3-ConvNeXt), we used a SegFormer-style MLP decoder~\citep{xie_segformer_2021}, which projects each scale to a common width, upsamples to the finest scale, and fuses them. DiceCE loss (equal weighting for Dice and cross-entropy) was used for all tasks. The full decoder specifications are included in Appendix~\ref{app:head-architectures}.

\section{Results}

\begin{figure*}[ht]
    \centering
    \includegraphics[width=\linewidth]{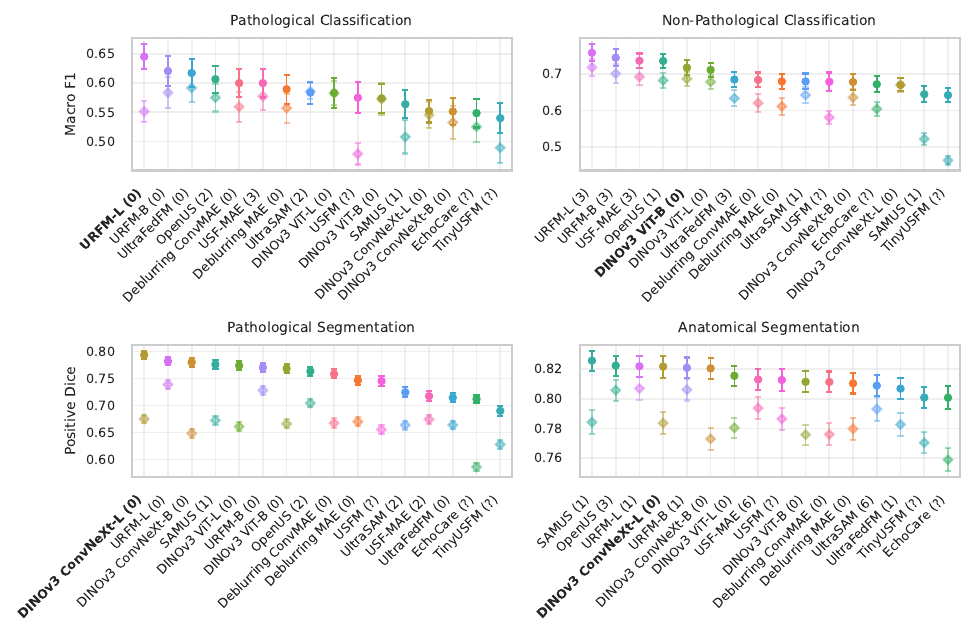}
    \caption{Full fine-tuning  (\markercircle) and head-only (\markerdiamond) performance in each category. The number in parentheses denotes the number of tasks whose data overlaps with the pretraining datasets (``?'' if the pretraining datasets are not disclosed). The \textbf{bold} model is the best-performing model with no overlap.}
    \label{fig:per_category_results}
\end{figure*}

Figure \ref{fig:per_category_results} shows the performance of each model in each category, under full fine-tuning and head-only training. The breakdown of performance on each task is included in Appendix \ref{app:per-task-results}. In Appendix~\ref{app:rank-stability}, we analyze rank stability and show that the tasks with smaller datasets do not unduly influence model rankings.

\paragraph{Anatomical vs. pathological segmentation.} All models struggle to segment pathological structures (e.g., tumors) more than anatomical ones (e.g., organs). The worst model's anatomical score (EchoCare, $0.801 \pm 0.008$) exceeds the best model's pathological score (DINOv3 ConvNeXt-L, $0.794 \pm 0.008$). Despite relatively high overall performance, some anatomical segmentation tasks remain very challenging. In particular, for kidney region segmentation all models achieve Dice scores  $\leq 0.487$.

\begin{figure*}
    \centering
    \includegraphics[width=0.8\linewidth]{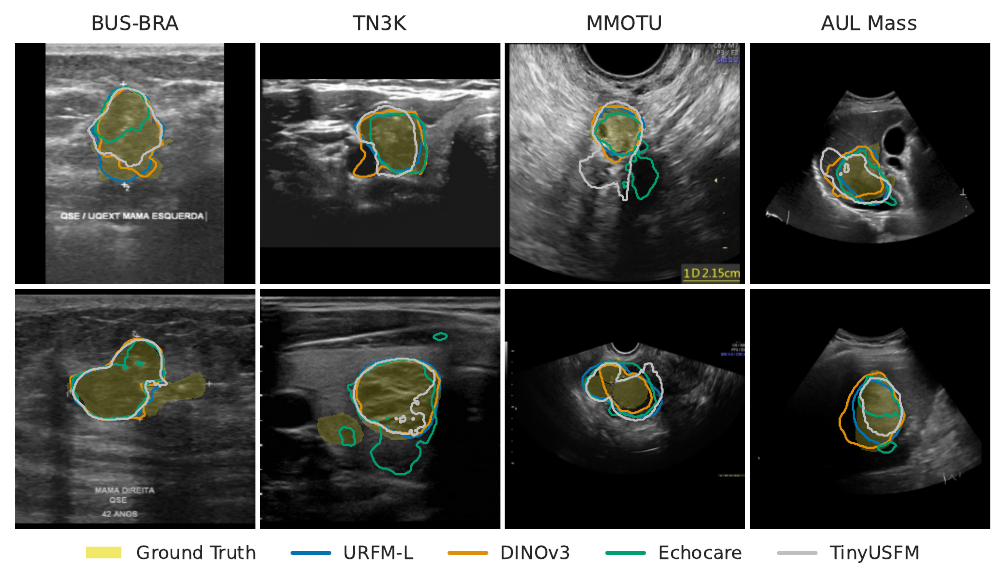}
    \caption{Ground truth and predicted segmentation boundaries for the two best and two worst pathological segmentation models.}
    \label{fig:pathological_segmentation_predictions}
\end{figure*}

To put the quantitative results in perspective, Figure~\ref{fig:pathological_segmentation_predictions}  presents qualitative examples for each pathological segmentation task. These illustrate both the remaining room for improvement among the best-performing models and the gap between the best and worst models. Additional examples for pathological and anatomical segmentation are provided in Appendix~\ref{app:segmentation-examples}.

\paragraph{Non-pathological vs. pathological classification.} As with segmentation, we find that pathological classification scores are consistently lower than the non-pathological  scores for all models. Here, the worst model's non-pathological score (TinyUSFM, $0.641 \pm 0.018$) is similar to the best model's pathological score (URFM-L, $0.645 \pm 0.021$).

\paragraph{Utility of domain-specific pretraining.} A surprising result is that the general-purpose DINOv3 ConvNeXt models rank among the best for segmentation tasks. DINOv3 ConvNeXt-L tops the leaderboard in pathological segmentation ($0.794\pm0.008$). It is also fourth overall in anatomical segmentation ($0.822 \pm 0.007$),  within the same tier as the top performing model, SAMUS ($0.826\pm0.007$), whose pretraining dataset overlaps with one of the tasks.

For non-pathological classification, DINOv3 ViT-B ($0.716 \pm 0.021$) is fifth overall and is the best performing model without pretraining overlap. However, the best performing DINOv3 model (ViT-L, $0.583 \pm 0.025$) falls much further from the lead in the more challenging pathological classification category led by URFM-L ($0.645 \pm 0.021$).

\paragraph{Head-only training vs. full fine-tuning.} The fully fine-tuned models consistently perform as well as, and usually much better, than their frozen counterparts across all categories. While head-only training is interesting for evaluating the quality of the representations learned by the models, at the scale of existing models (up to 307 million parameters), we recommend full fine-tuning to maximize performance. All of our training runs used less than \qty{20}{\giga\byte} of VRAM, which could have been further reduced using memory-saving techniques such as gradient accumulation.

\paragraph{Model size vs. performance.} Of the models that are available in multiple sizes (URFM, DINOv3 ViT, and DINOv3 ConvNeXt), the larger variants typically score higher in each category than the smaller ones, but the differences are usually not significant. TinyUSFM, the smallest model, with only \qty{5.5}{\million} parameters, is among the worst performing models in each category. Figure \ref{fig:performance_vs_size} in Appendix \ref{app:performance_trends} visualizes the relationship between performance and size.

\paragraph{The evolution of performance over time.} Comparing the performance and release dates of the models, we find that many models do not outperform their predecessors within our setup. For example, the URFM and DINOv3 models (released in August 2025) and SAMUS (released in September 2023) all outperform subsequently released models, such as USF-MAE, EchoCare, and TinyUSFM. Figures that illustrate performance over time are included in Appendix \ref{app:performance_trends}.

\section{Discussion}

\subsection{Recommendations for Future Models}

Our initial benchmark scores demonstrate the relative weakness of current models on pathological tasks, which should be a focus for future models. They also highlight several potential avenues for improvement.

Models trained via knowledge distillation (URFM, DINOv3) perform very well, with the exception of TinyUSFM, a distilled version of USFM. In particular, URFM's approach of distilling knowledge from the multi-domain BiomedCLIP model appears to be a key differentiating factor in its leading pathological classification performance, given that its architecture (ViT) and training data scale (1,003,465 images) are not outliers. URFM also leads on non-pathological classification, though both the CAMUS and Fetal Planes datasets that the tasks are derived from are part of its pretraining corpus, so that result should be interpreted with caution.

Larger pretraining datasets and greater diversity do not necessarily result in better models. Despite having some of the largest pretraining datasets of any ultrasound models, USFM and EchoCare perform poorly relative to other models. The DeblurringMIM models, pretrained exclusively on private thyroid ultrasound images, match or exceed both models in all categories. One key difference may be data quality: URFM and the DeblurringMIM models are trained partially or exclusively, respectively, on private data. This may be higher quality than data collated from public sources.

Modern CNN architectures and pretraining methods may warrant revisiting. The strong performance of DINOv3 models, particularly for segmentation, suggests that certain combinations of architecture (e.g., ConvNeXt + SegFormer), pretraining method, and pretraining scale for general-purpose vision models have overcome the benefits of domain-specific pretraining with simpler approaches. We observed that the performance gains from the SegFormer decoder are largely limited to head-only training in our ablations (see Appendix~\ref{app:segmentation-decoder-ablations}), but the impact of backbone and decoder architectures should be more deeply investigated to isolate their effects. The best ultrasound foundation models retain their segmentation advantage only when the backbone is frozen. We hypothesize that this reflects domain shift between DINOv3's pretraining data and the test data.

\subsection{Limitations}

Several limitations warrant discussion. First, the benchmark is built on public datasets. While this problem is not unique to UltraBench 2, it means that we rely on collaboration from the community to exclude these datasets during pretraining to keep comparisons meaningful. Sourcing diverse, private evaluation data at this scale remains a broader challenge for the field as a whole.

Second, the hyperparameters, augmentations, and task heads were not optimized per model. This was infeasible given the scale of our evaluations and available resources. As such, our results should be viewed as a foundation from which to build on. We invite the community to explore these directions and submit improved results for new and existing models alike.

Finally, we hope that UltraBench 2 will serve the community well for years to come, but we recognize that all benchmarks reach saturation. We intend for the benchmark to evolve (e.g., by adding new tasks, resolutions, etc.) to combat this and meet the evolving needs of the community.

\section{Conclusions}

Through UltraBench 2, we provide a standardized evaluation of existing ultrasound foundation models and highlight their relative strengths and weaknesses against one another and against general-purpose models. Our results show that pathological tasks remain the greatest weakness across models, and that several newer models fail to outperform older ones under our protocol. This underscores the need for standardized, reproducible evaluation to separate genuine capability gains from artifacts of inconsistent evaluations. We hope UltraBench 2 offers a more complete picture of where ultrasound foundation models stand today and that the community adopts this benchmark as a common performance standard in future work.

\acks{This research was enabled in part by support provided by Calcul Qu\'ebec (\href{https://www.calculquebec.ca}{calculquebec.ca}) and the Digital Research Alliance of Canada (\href{https://alliancecan.ca/}{alliancecan.ca}). It was also supported through funding from the Canadian Institute for Advanced Research (\href{https://cifar.ca/}{cifar.ca}) and the Natural Sciences and Engineering Research Council of Canada (\href{https://www.nserc-crsng.gc.ca/}{nserc-crsng.gc.ca}). This work was also supported by the UCLA Radiology Exploratory Research Grant.}

\bibliography{references}

\clearpage
\appendix
\let\oldsection\section
\renewcommand{\section}{\clearpage\oldsection}
\setcounter{dblbotnumber}{3}            % up to 3 wide floats at a page bottom
\setcounter{dbltopnumber}{3}            % up to 3 wide floats at a page top
\renewcommand{\dbltopfraction}{0.9}
\renewcommand{\dblfloatpagefraction}{0.8}

\section{Additional Information on Ultrasound Foundation Models}
\label{app:ultrasound-foundation-models}

\subsection{USFM}

USFM \citep{jiao2024} introduced a spatial-frequency dual masking scheme built on a masked image modeling (MIM) backbone to improve representational robustness from low-quality ultrasound images. Pretraining was conducted on the 3M-US database, the first multi-organ, multi-center, and multi-device ultrasound collection. This database had over two million images from 12 human organs. Evaluation was conducted via full fine-tuning of USFM as a pretrained backbone encoder for segmentation and classification tasks. Downstream experiments spanned three task types: segmentation of key structures (brachial plexus, thyroid nodule, breast tumor, and fetal abdomen), image classification (breast tumor malignancy, multi-organ presence, and fetal planes) and image enhancement. We evaluate the released ViT-B encoder with \qty{85.7}{\million} parameters.

\subsection{TinyUSFM}

TinyUSFM \citep{ma2026} aims to reduce the computational cost of ultrasound foundation models by distilling USFM into a \qty{5.5}{\million}-parameter ViT-Tiny model. The resulting student, with roughly 6.4\% of the teacher's parameters, was modified to retain the same number of attention heads as the teacher so that features could be aligned. The framework has two components: 1) feature-gradient driven coreset selection is used to develop a diverse 200,000-image training set from the 2.1 million images of USFM's 3M-US pretraining dataset and 2) a domain-separated masked image modeling strategy constructs two masked views of each image, perturbing the images in spatial and frequency domains, and scales the feature distillation loss by the cosine similarity between the teacher's representations of the two views, encouraging knowledge transfer where the teacher is consistent across corruptions. Evaluation was performed with a full fine-tuning setup using a single linear layer head for classification and a lightweight pyramid neck with an FPN-style decoder for segmentation. Evaluation was performed on UniUS-Bench, a dataset benchmark assembled by the authors from public data comprising 8 classification and 10 segmentation tasks across 15 organs. This is the smallest model we evaluated by a wide margin.

\subsection{URFM}

URFM \citep{kang2025} proposes a representation-based masked image modeling framework that has the ViT encoder reconstruct high-level semantic features provided by BiomedCLIP, avoiding a potentially inefficient pixel-reconstruction objective \citep{ivezic2026} and addressing the low signal-to-noise ratio of ultrasound. The pretraining dataset had over one million ultrasound images (approximately 50\% was proprietary) spanning 15 major anatomical organs. Evaluation was based on full fine-tuning the backbone encoder for classification tasks. Ten downstream ultrasound clinical applications were evaluated across nine organs. We evaluated both released sizes: the ViT-B encoder with \qty{85.8}{\million} parameters and the ViT-L encoder with \qty{303.3}{\million}.

\subsection{USF-MAE}

USF-MAE \citep{megahed2026} follows a standard masked autoencoder (MAE) framework adapted for ultrasound, using a ViT backbone as the encoder and a lightweight transformer decoder for image reconstruction from masked patches. The model was pretrained on the OpenUS-46 dataset, which has about 370,000 2D and 3D ultrasound images from 46 open-source datasets. Evaluation was performed exclusively via fine-tuning of the pretrained encoder for three downstream classification datasets: BUS-BRA (breast cancer), MMOTU-2D (ovarian tumors), and GIST514-DB (gastrointestinal stromal tumors). We evaluated the ViT-B encoder with \qty{85.8}{\million} parameters.

\subsection{EchoCare}

EchoCare \citep{zhang2025} extends the standard self-supervised MAE architecture with a novel dual-branch design: an image encoder and image decoder perform pixel-level masked reconstruction, while a second anatomy-classifier decoder jointly learns hierarchical anatomical relationships. Pretraining was conducted on EchoCareData, a public dataset of 4.5 million ultrasound images encompassing 9 major body regions and 52 anatomical organs. Evaluation was performed via full fine-tuning the backbone encoder for 10 tasks in a downstream suite spanning disease diagnosis classification, lesion segmentation, organ detection, fetal brain landmark prediction, cardiac ejection fraction regression, image quality enhancement, and report generation. We evaluate the Swin Transformer encoder, which has \qty{120.2}{\million} parameters.

\subsection{OpenUS}

OpenUS \citep{zheng2025} is an ultrasound foundation model built exclusively on publicly available data. OpenUS uses a Vision Mamba (VMamba) backbone, which captures both local and long-range spatial dependencies via selective state-space modeling. A key novelty is the self-adaptive masking strategy during training, which combines a teacher attention map with a student reconstruction loss into a unified Adaptive Learning Priority (ALP) score, directing the masking toward the most clinically informative regions. Pretraining was conducted on 308,584 images from 38 publicly available ultrasound datasets covering 12 organs. Evaluation was performed via fine-tuning a task head (keeping the encoder frozen) for two downstream classification datasets (BUSI breast cancer and Fetal Planes) and two segmentation datasets (BUS-BRA breast lesions and TN3K thyroid nodules). We evaluated the \qty{49.4}{\million}-parameter VMamba encoder, the smallest backbone after TinyUSFM.

\subsection{UltraFedFM}

UltraFedFM \citep{jiang2025} was first released as a preprint in November 2024. The main contribution is a federated pretraining framework. Sixteen geographically distinct clients each contributed to pretraining the model (a masked autoencoder) on their private ultrasound data, only sharing their updated model parameters between rounds for aggregation. To handle cross-site image differences, the framework introduces structured masking with adaptive thresholding (SMAT), which adjusts masking patterns to organ- and lesion-specific texture features, and a multi-scale image corruption (MIC) branch for improving robustness to low-quality ultrasound inputs. The pretraining dataset had a combined 1,015,754 unlabeled images covering 19 organs. Evaluation was performed via full fine-tuning of the encoder for disease diagnosis tasks across 8 kinds of organs and 6 modalities and lesion segmentation across 5 datasets. This paper also conducts a clinical comparison on 8 systemic malignant diseases, demonstrating that UltraFedFM outperforms mid-level sonographers (4--8 years of experience) and matches expert-level sonographers ($>$10 years of experience). We evaluated the ViT-B encoder, which has \qty{85.8}{\million} parameters.

\subsection{DeblurringMIM}

DeblurringMIM \citep{kang2024a} was one of the first applications of masked image modeling to ultrasound. The work was motivated by the observation that standard masked autoencoder pretraining failed to account for ultrasound's low signal-to-noise ratio and the importance of fine-grained diagnostic details. DeblurringMIM extends the MAE objective by adding a blurring step, where inputs are first blurred using one of several blur operators before random masking. Then, the model is trained to reconstruct the original sharp image from the masked and blurred input. The encoder is a multi-scale hybrid convolution-transformer based on ConvMAE. Pretraining was conducted on approximately 280,000 thyroid ultrasound images. Downstream evaluation is performed with full fine-tuning on tasks spanning classification of thyroid nodules and Hashimoto's thyroiditis, and segmentation of thyroid nodules. We evaluated both released variants: the \qty{85.8}{\million}-parameter ViT-B encoder and the hybrid \qty{88.9}{\million}-parameter ConvMAE encoder.

\subsection{SAMUS}

SAMUS \citep{lin2024c} adapts Segment Anything Model (SAM) for ultrasound segmentation by introducing a parallel CNN branch alongside SAM's ViT encoder, injecting local feature information via cross-branch attention, and further incorporating a position adapter and feature adapter to retune the model from large-scale natural image inputs (1024$\times$1024 px) to more practical input sizes (256$\times$256 px). The model was trained and evaluated on an ultrasound dataset of approximately 30,000 images and 69,000 masks covering six anatomical object categories from publicly available sources. Downstream evaluation was conducted via full fine-tuning. Tasks were focused exclusively on interactive and automatic segmentation across the six ultrasound object categories in the US30K dataset. We evaluated the full adapted encoder with \qty{129.3}{\million} parameters.

\subsection{UltraSam}

UltraSam \citep{meyer2025} was first released as a preprint in November 2024. The framework fully fine-tunes the Segment Anything Model (SAM) on a large ultrasound dataset under a prompt-conditioned segmentation setup, supporting both point- and bounding-box prompts. Pretraining data consisted of US-43d, a compilation of 43 open-access ultrasound segmentation datasets totaling over 280,000 image--mask pairs across 20 diverse clinical applications. Evaluation employed two protocols: prompt-based zero-shot segmentation benchmarked on three diverse public datasets, and downstream fine-tuning of an UltraSam-initialized Vision Transformer. Downstream tasks spanned both interactive segmentation of anatomical structures and image-level classification. We evaluated the fine-tuned SAM ViT encoder with \qty{85.9}{\million} parameters.

\subsection{DINOv3}

DINOv3 \citep{simeoni2025} extends DINOv2's self-supervised student-teacher distillation approach to significantly larger model scales. One of the main contributions is Gram anchoring, a training technique that regularizes cosine similarity between student and teacher patch features using a Gram-matrix objective. This reduces the degradation of dense feature maps that emerges in larger models trained for long schedules. The initial 7B model is  distilled into a family of smaller variants (ViT-Small, ViT-Base, ViT-Large, and ConvNeXt-based architectures). Pretraining was conducted on a curated set of 1.689 billion images collected from public posts on Instagram (LVD-1689M). DINOv3 is included in our evaluation as a strong general-purpose vision foundation model baseline, assessed via the standard probing and fine-tuning protocol used for the other models. We evaluated four distilled variants: ViT-B/16 (\qty{86}{\million} parameters) and ViT-L/16 (\qty{307}{\million} parameters), and ConvNeXt-Base (\qty{89}{\million} parameters) and ConvNeXt-Large (\qty{198}{\million} parameters).

\section{Dataset Details}
\label{app:datasets}

\begin{figure*}[!htbp]
    \centering
    \includegraphics[width=\linewidth]{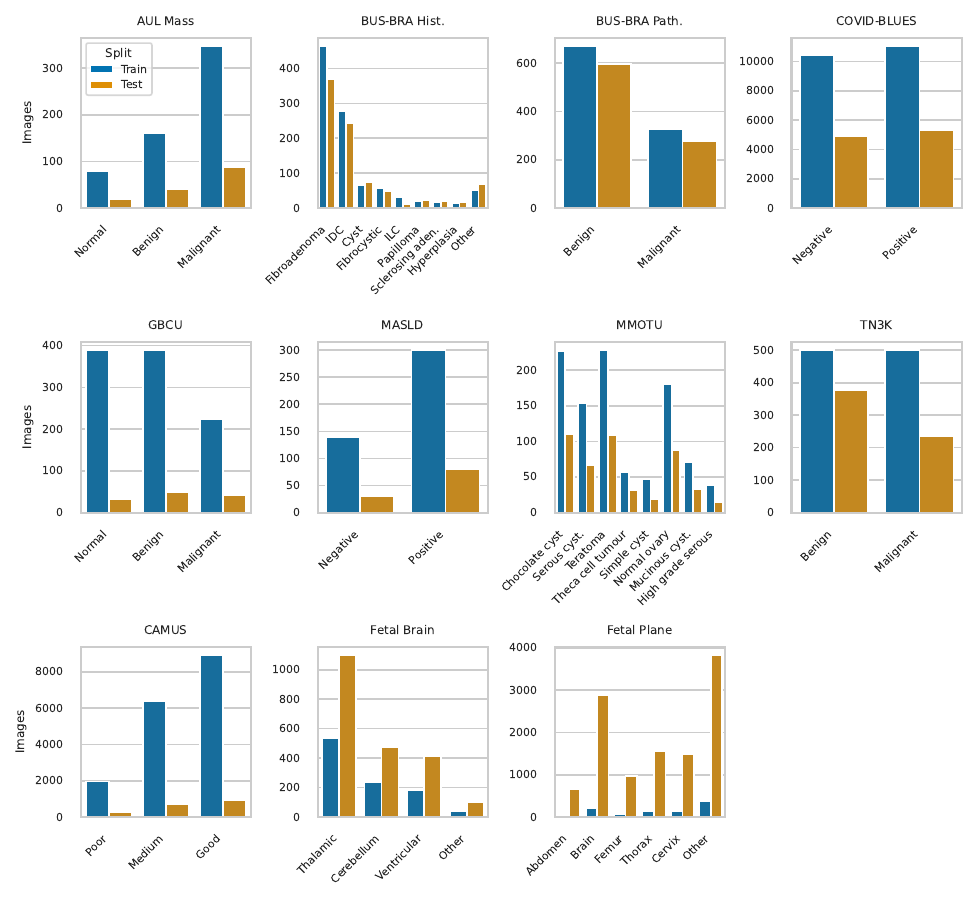}
    \caption{The class distributions of the training and test splits for each classification task.}
    \label{fig:classification_class_distributions}
\end{figure*}

\begin{figure*}[!htbp]
    \centering
    \includegraphics[width=\linewidth]{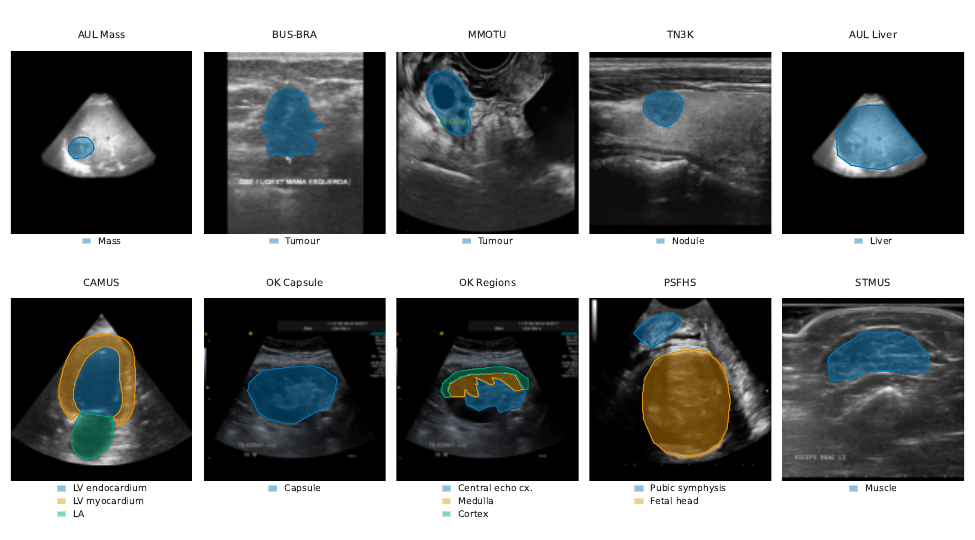}
    \caption{Examples of the images and segmentation masks for each segmentation task.}
    \label{fig:segmentation_masks}
\end{figure*}

Here we detail the origin, licensing, and processing of every dataset in UltraBench~2. Dataset-level details are reported in Table~\ref{tab:tasks}. Class-level label distributions and segmentation mask examples are included in Figures~\ref{fig:classification_class_distributions} and~\ref{fig:segmentation_masks}.

\subsection{Common Processing Pipeline}
\label{app:datasets-pipeline}

All datasets are converted to a single Hugging Face \texttt{DatasetDict}
format so that the training and evaluation layers are dataset-agnostic. The
following steps are applied uniformly.

\paragraph{Splits.} Every dataset is published with a \texttt{train} split and a \texttt{test} split. Whenever patient identifiers are present, splitting is performed at the patient level so that there is no patient-level leakage between development and evaluation. In our experiments, we used a validation set for hyperparameter tuning and early stopping that was an 80:20 split of \texttt{train}, grouped by patient wherever patient identifiers are available and stratified by the task label where the class distribution is otherwise unstable. The \texttt{test} split is never used during model selection.

\subsection{Dataset Summary}
\label{app:datasets-summary}

Table~\ref{tab:dataset-summary} summarizes the licensing and access terms
for each dataset. The paragraphs below describe each dataset.

\paragraph{AUL.} The Annotated Ultrasound Liver dataset \citep{xu_improving_2023} comprises 735 liver ultrasound images with liver and mass segmentation masks and a three-class mass label (normal, benign, malignant). We apply a label-stratified 80:20 split. One correction is noted -- image \texttt{374.jpg} has no liver mask and is therefore excluded from the liver segmentation task, giving that task one fewer training frame (587) than the mass tasks (588). Both segmentation masks and the mass label were produced by two radiologists with more than ten years of experience, one annotating and the other verifying, with disagreements resolved by consensus. Malignancy is established by biopsy or post-surgical pathology where tissue was obtained, and otherwise by contrast-enhanced imaging in patients with hepatitis~B- or C-related cirrhosis or a known primary cancer elsewhere, so the label is not uniformly pathology-confirmed.

\paragraph{BUS-BRA.} BUS-BRA \citep{gomez-flores_bus-bra_2024} contains 1,875 breast ultrasound images from 1,064 patients with biopsy-proven tumor labels, BI-RADS categorization, and tumor segmentation masks. The split is performed by patient. The dataset supports three tasks: tumor segmentation, binary pathology classification, and nine-class histology classification. The benign and malignant labels and the histopathological types are biopsy-proven. A senior ultrasonographer manually determined lesion outlines and assigned the BI-RADS categories.

\paragraph{CAMUS.} CAMUS \citep{leclerc_deep_2019} is a two-dimensional echocardiography dataset of 500 patients with expert-derived cardiac structure annotations and sequence-level image-quality ratings. We use the original train/validation/test split and extract frames and masks from the released sequences. The dataset is published as two subsets. Image quality is a sequence-level label, and the training split contains 900 sequences, so the classification subset retains all frames. The segmentation subset is capped at 1,000 frames drawn uniformly at random within each (patient, view) sequence, which promotes coverage of every training patient and both views. Contours for all 500 patients were traced by a single expert cardiologist, who also assigned the image-quality grade.

\paragraph{COVID-BLUES.} COVID-BLUES \citep{wiedemann_covid-blues_2025} is a lung ultrasound dataset in which each patient is imaged at standardized BLUE-protocol positions. We split the dataset by patient, assigning 43 patients to training and 20 to test. The 63 patients contribute a total of 362 videos containing 31,746 frames. The label is the patient's RT-PCR test result rather than an image-level annotation; the label applies to every frame acquired from that patient, which is why the task is scored per patient.

\paragraph{Fetal Planes.} The Fetal Planes dataset \citep{burgos-artizzu_evaluation_2020} contains 12,400 maternal-fetal ultrasound images from 1,792 patients labeled by anatomical plane. We preserve the original train/test split and derive two subsets: \texttt{fetal}, labeled by fetal plane (six classes: abdomen, brain, femur, thorax, maternal cervix, and other), and \texttt{fetal\_brain}, restricted to brain images and labeled by brain plane (four classes: trans-thalamic, trans-cerebellum, trans-ventricular, and other). Each subset is capped independently at 1,000 training images by moving whole patients into the corresponding test split. A patient may therefore appear in the training split of one subset and the test split of the other, so the two subsets should not be pooled to avoid patient-level leakage. All images were classified manually by a single senior maternal-fetal specialist, who also assigned the brain sub-plane labels.

\paragraph{GBCU.} The Gallbladder Cancer Ultrasound dataset \citep{basu_surpassing_2022} provides gallbladder ultrasound images labeled normal, benign, or malignant. We use the predefined split and cap the training split at 1,000 images by removing frames from the majority classes first. Patient identifiers are not publicly available, so surplus frames are discarded rather than moved to the test split, to avoid unintentionally introducing patient-level leakage. The three-class labels are biopsy-proven.

\paragraph{MASLD.} The MASLD dataset \citep{byra_transfer_2018} comprises 550 liver ultrasound images from 55 patients with a binary label indicating a diagnosis with Metabolic dysfunction-associated steatotic liver disease (MASLD), formerly known as nonalcoholic fatty liver disease (NAFLD). An 80:20 split is performed at the patient level to establish train and test splits. The level of steatosis was graded from a wedge liver biopsy by a single pathologist, and a patient is counted as positive when more than 5\% of hepatocytes show fatty infiltration. Every frame from a patient inherits the assigned label.

\paragraph{MMOTU-2D.} MMOTU-2D \citep{zhao_mmotu_2023} contains 1,469 ovarian tumor ultrasound images with eight-class tumor-type labels and binary tumor segmentation masks. The eight tumor types (chocolate cyst, serous cystadenoma, teratoma, theca cell tumor, simple cyst, normal ovary, mucinous cystadenoma, and high-grade serous cystadenocarcinoma) are determined through histopathological analysis. The dataset is distributed according to the original train and test splits.

\paragraph{Open Kidney.} The Open Kidney dataset \citep{singla_open_2023} provides 514 kidney ultrasound images with capsule and multi-region annotations (background, central echo complex, medulla, and cortex). We apply an 80:20 split stratified by view label (longitudinal, transverse, other). Each image corresponds to a distinct patient, so patient overlap is not a concern during splitting. Each image was annotated independently by two registered sonographers, each with at least 30 years of experience, following a protocol agreed in advance. Two biomedical engineers then reviewed every annotation. The two annotation sets were released separately rather than fused, and we use those of the first sonographer throughout.

\paragraph{PSFHS.} PSFHS \citep{chen_psfhs_2024} contains 1,358 intrapartum ultrasound images from 1,124 patients annotated for pubic symphysis and fetal head segmentation. We assign 1,000 images to the training split and the remainder to test. Other than the total number of patients, the original dataset did not provide patient metadata, so it is not possible to guarantee the absence of patient overlap between the splits. Masks were produced by a team of 2 physicians and 18 biomedical students, all trained beforehand, with each image annotated by two annotators and disputed pixels reviewed and corrected by a physician with a decade of experience.

\paragraph{STMUS.} The transverse musculoskeletal ultrasound dataset \citep{marzola_deep_2021} consists of 8,169 images, with segmentation masks for three different muscles, the tibialis anterior, gastrocnemius medialis, and biceps brachii. The original data release included a patient characteristics metadata file that allowed us to match images to corresponding patients. The dataset is then split 80:20 by patient and the training split is capped at 1,000 frames by moving whole patients into the test split. The masks are manual segmentations by a neurodiagnostic technician, with 5--10+ years of experience in quantitative muscle imaging.

\paragraph{TN3K.} TN3K \citep{gong_thyroid_2023} provides thyroid ultrasound images with nodule segmentation masks and binary malignancy labels. The dataset is divided according to the original train and test splits of 2,879 training and 614 test images. We preserve the split and cap the training set at 1,000 images, removing frames from the majority classes first so that the rarer labeled nodules are retained. The surplus frames are discarded because the images do not have any patient identifiers. The nodule masks and the benign and malignant labels are attributed to experienced radiologists.

\subsection{Licensing and Access}
\label{app:datasets-licensing}

The benchmark redistributes prepared versions of these datasets only where the original license allows it. Two datasets, GBCU and Open Kidney, are governed by research usage agreements and cannot be freely redistributed. The COVID-BLUES and CAMUS datasets have licenses that restrict commercial use. For every dataset, we ship download and processing scripts that reconstruct the benchmark version from the original source, so the full benchmark remains reproducible by any user who obtains access through the original terms. Users of UltraBench~2 are asked to cite the original dataset publications listed in Table~\ref{tab:dataset-summary} alongside this work.

\subsection{Standardizing Anatomical Region Counts}
\label{app:anatomical-regions}

Prior benchmarks report anatomical region counts using inconsistent conventions. Some treat every named sub-structure as a distinct region, while others report an aggregate count that includes pathological targets as well as the corresponding organs. To enable a fair comparison in Table~\ref{tab:benchmarks}, we did not use each paper's self-reported count. Instead, we manually reviewed the dataset-level descriptions in each benchmark and labeled every dataset using a fixed set of anatomical regions (e.g., breast, liver, kidney, heart, thyroid), grouping closely related structures under a single label (e.g., carotid artery and general cardiac imaging were both grouped under \textit{cardiovascular} and multiple muscle-specific datasets were grouped under \textit{muscle}). Datasets that did not specify individual organs, or that spanned several  adjacent anatomies under one label (e.g., a general abdominal ultrasound dataset without organ-level information), were assigned to a broader \textit{abdomen} category rather than being split arbitrarily. Under these universal definitions, we counted the number of unique labels present in each benchmark to standardize this tally across papers.

\begin{table*}[!htbp]
  \centering
  \caption{Origin and licensing of the UltraBench~2 datasets. ``Access'' indicates whether the prepared version is distributed directly through our Hugging Face organization, or must instead be obtained from the original provider and then processed with UltraBench 2. Task-level frame counts are reported in Table~\ref{tab:tasks}.}
  \label{tab:dataset-summary}
  \small
  \setlength{\tabcolsep}{6pt}
  \begin{threeparttable}
  \begin{tabular}{@{}lll@{}}
    \toprule
    Dataset & Access & License \\
    \midrule
    AUL \citep{xu_improving_2023}                             & Hugging Face                & CC BY 4.0 \\
    BUS-BRA \citep{gomez-flores_bus-bra_2024}             & Hugging Face                & CC BY 4.0 \\
    CAMUS \citep{leclerc_deep_2019}                    & Hugging Face                & CC BY-NC-SA 4.0 \\
    COVID-BLUES \citep{wiedemann_covid-blues_2025}       & Hugging Face                & CC BY-NC-ND 4.0 \\
    Fetal Planes \citep{burgos-artizzu_evaluation_2020} & Hugging Face                & CC BY 4.0 \\
    GBCU \citep{basu_surpassing_2022}                         & Original provider\tnote{a}  & Research use \\
    MASLD \citep{byra_transfer_2018}                       & Hugging Face                & CC BY 4.0 \\
    MMOTU-2D \citep{zhao_mmotu_2023}                     & Hugging Face                & CC BY-NC-ND 4.0 \\
    Open Kidney \citep{singla_open_2023}          & Original provider\tnote{b}  & Research use \\
    PSFHS \citep{chen_psfhs_2024}                       & Hugging Face                & CC BY 4.0 \\
    STMUS \citep{marzola_deep_2021}                    & Hugging Face                & CC BY 4.0 \\
    TN3K \citep{gong_thyroid_2023}                         & Hugging Face                & MIT \\
    \bottomrule
  \end{tabular}
  \begin{tablenotes}[flushleft]
  \footnotesize
  \item[a] Managed through a research usage agreement that prohibits
    redistribution. Request access at \url{https://gbc-iitd.github.io/data/gbcu}.
  \item[b] Managed through a research usage agreement that prohibits
    redistribution. Request access at \url{https://rsingla.ca/kidneyUS/}.
  \end{tablenotes}
  \end{threeparttable}
\end{table*}

\section{Training Setup}
\label{app:training-setup}

\paragraph{Head-only training.} The backbone was frozen and set to evaluation mode. Normalization statistics and any stochastic components (dropout, etc.) were also frozen. No augmentation was applied. With the frozen backbone we found that augmentation only harmed performance.

\paragraph{Full fine-tuning.} For the first 1,000 steps, training was identical to head-only training, except that augmentation was used from step 0. The backbone was unfrozen at step 1,000 (except for batch normalization layers, where applicable) and the parameters were partitioned into $G=4$ contiguous groups by parameter registration order (shallowest to deepest). This purely order-based grouping requires no architecture-specific layer naming, so it applies uniformly to all ViT, ConvNeXT, and CNN backbones. We used layer-wise learning rate decay to adjust the learning rate for each group:

$$
\mathrm{lr}_{\text{backbone}}(j,t) = \mathrm{lr}_{\text{head}}(t)\times r_{\max}\times d^{\,(G-1-j)}\times w(t)
$$

where $j=0$ is the shallowest group, $j=G-1$ the deepest; $r_{\max} = 0.01$ caps the deepest group's rate relative to the head's current rate; $d = 0.9$ discounts each successively shallower group by an additional factor of $d$; and

$$
w(t) = \mathrm{clip}\!\left(\frac{t-t_{\text{unfreeze}}}{\text{backbone\_warmup\_steps}},\,0,\,1\right)
$$

linearly ramps from 0 to 1 over the 1,000 steps immediately after unfreezing. Because the backbone rate is always a fixed proportion of the head's own decaying rate, it decays in lockstep with the head's cosine schedule for the rest of training rather than following an independent trajectory. Additional hyperparameters are detailed in Table~\ref{tab:hyperparams}.

\begin{table}[!htbp]
  \small
  \centering
  \caption{Optimization hyperparameters. These are used for both fine-tuning protocols unless noted.}
  \label{tab:hyperparams}
  \begin{tabular}{@{}ll@{}}
    \toprule
    Parameter & Value \\
    \midrule
    Optimizer              & AdamW \\
    Head LR                & $1\mathrm{e}{-}3$ \\
    Weight decay           & $1\mathrm{e}{-}2$ \\
    Schedule               & Cosine annealing ($\eta_{\min} = 0$) \\
    Max steps              & \num{10000} \\
    Batch size             & 32 \\
    Grad. accumulation     & 1 \\
    Max grad. norm         & 1.0 \\
    Val check interval     & 100 steps \\
    Patience               & 10 checks \\
    \bottomrule
  \end{tabular}
\end{table}

\section{Performance Trends}
\label{app:performance_trends}

This appendix relates overall performance to two factors: when a model was released, and how large it is. Figures~\ref{fig:classification_performance_over_time} and~\ref{fig:segmentation_performance_over_time} plot each model's mean score across all segmentation and all classification tasks against its release date. Figure~\ref{fig:performance_vs_size} places classification against segmentation performance with point size representing trainable parameter count, and shows that parameter count on its own does not lead to better performance.

\begin{figure*}[!bp]
\floatconts
  {fig:classification_performance_over_time}
  {\caption{Mean F1 score over all classification tasks for each model. The dashed line shows the evolution of the best model over time.}}
  {\includegraphics[width=\linewidth]{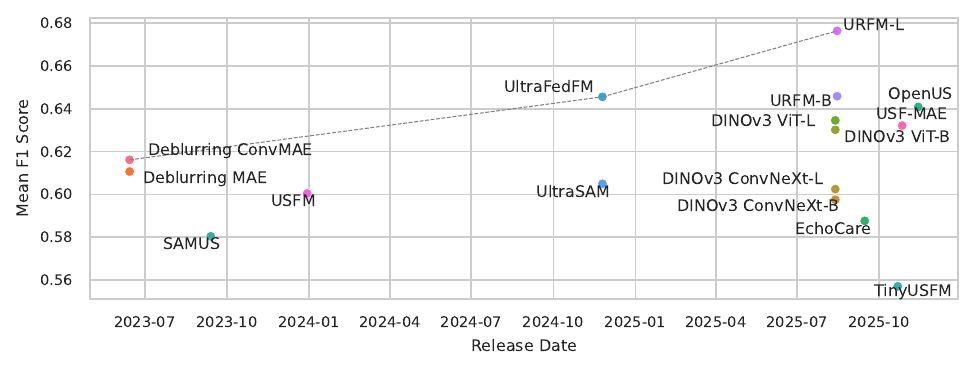}}
\end{figure*}

\begin{figure*}[!bp]
\floatconts
  {fig:segmentation_performance_over_time}
  {\caption{Mean positive Dice score over all segmentation tasks for each model. The dashed line shows the evolution of the best model over time.}}
  {\includegraphics[width=\linewidth]{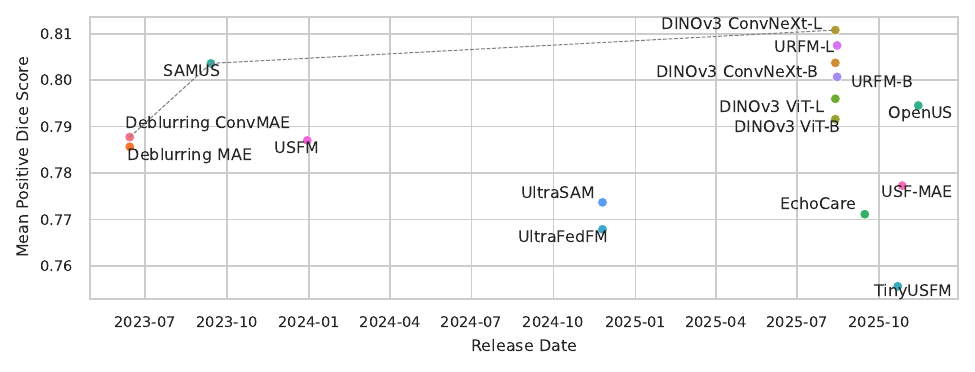}}
\end{figure*}

\begin{figure*}[!bp]
\floatconts
  {fig:performance_vs_size}
  {\caption{Overall classification and segmentation performance per model. The size of each point represents the number of trainable parameters.}}
  {\includegraphics[width=\linewidth]{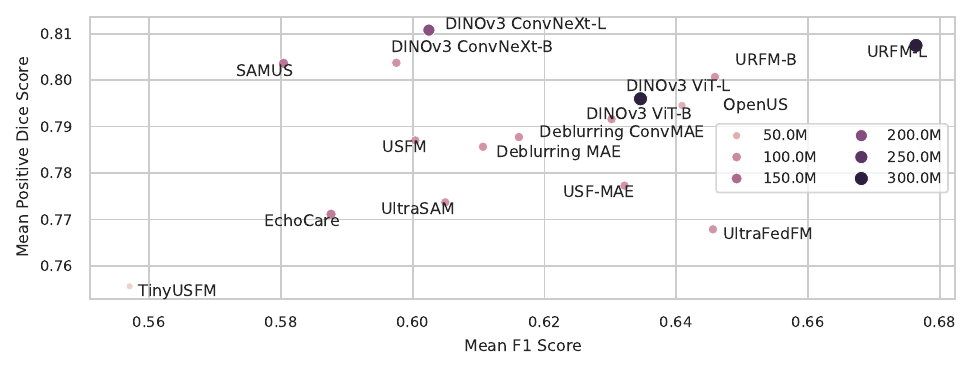}}
\end{figure*}

\section{Per-Task Results}
\label{app:per-task-results}

\begin{figure*}[!htbp]
    \centering
    \includegraphics[width=0.98\linewidth]{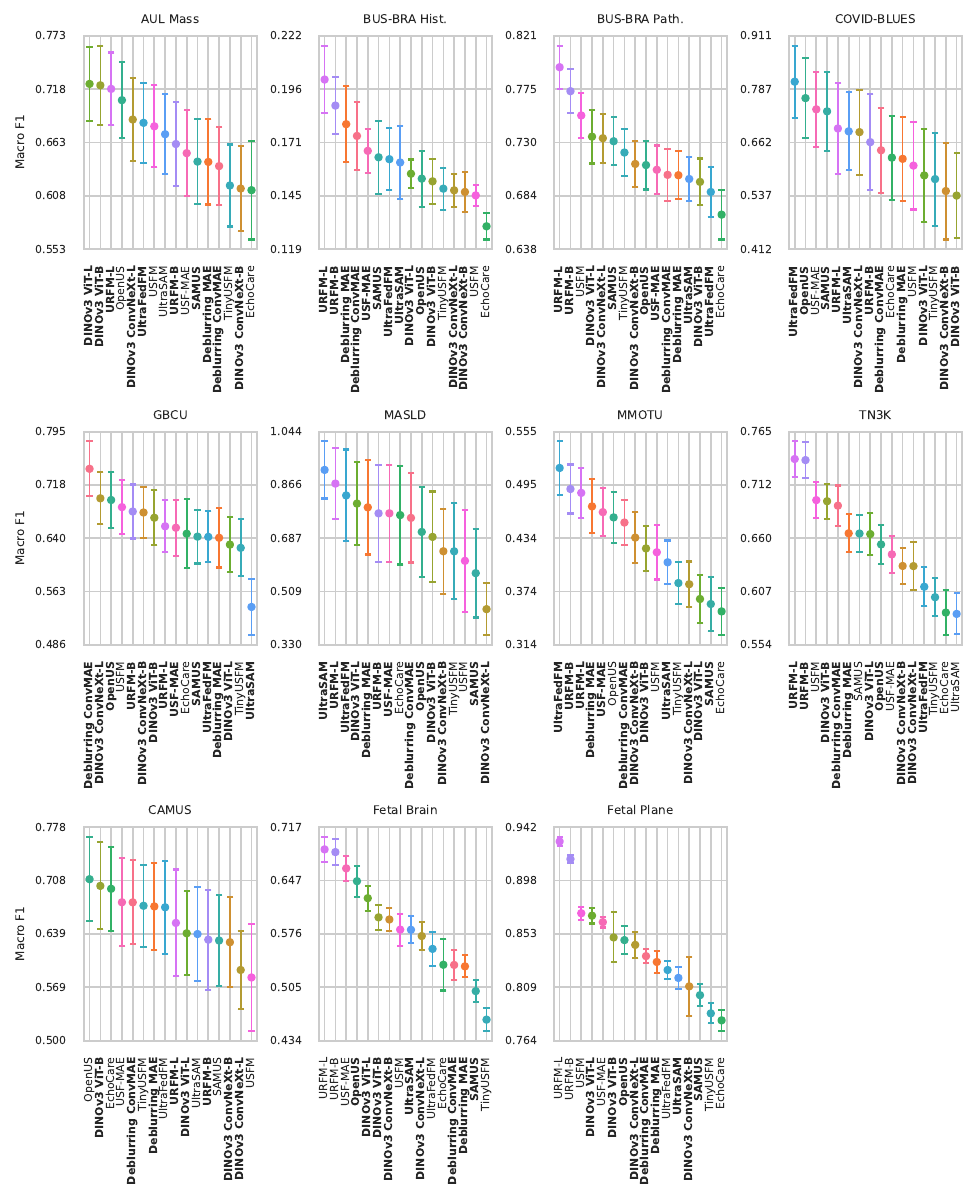}
    \caption{Full fine-tuning results on each classification task. \textbf{Bold} indicates that the task dataset was not included in the model's pretraining data.}
    \label{fig:per_task_classification_fft}
\end{figure*}

\begin{figure*}[!htbp]
    \centering
    \includegraphics[width=0.98\linewidth]{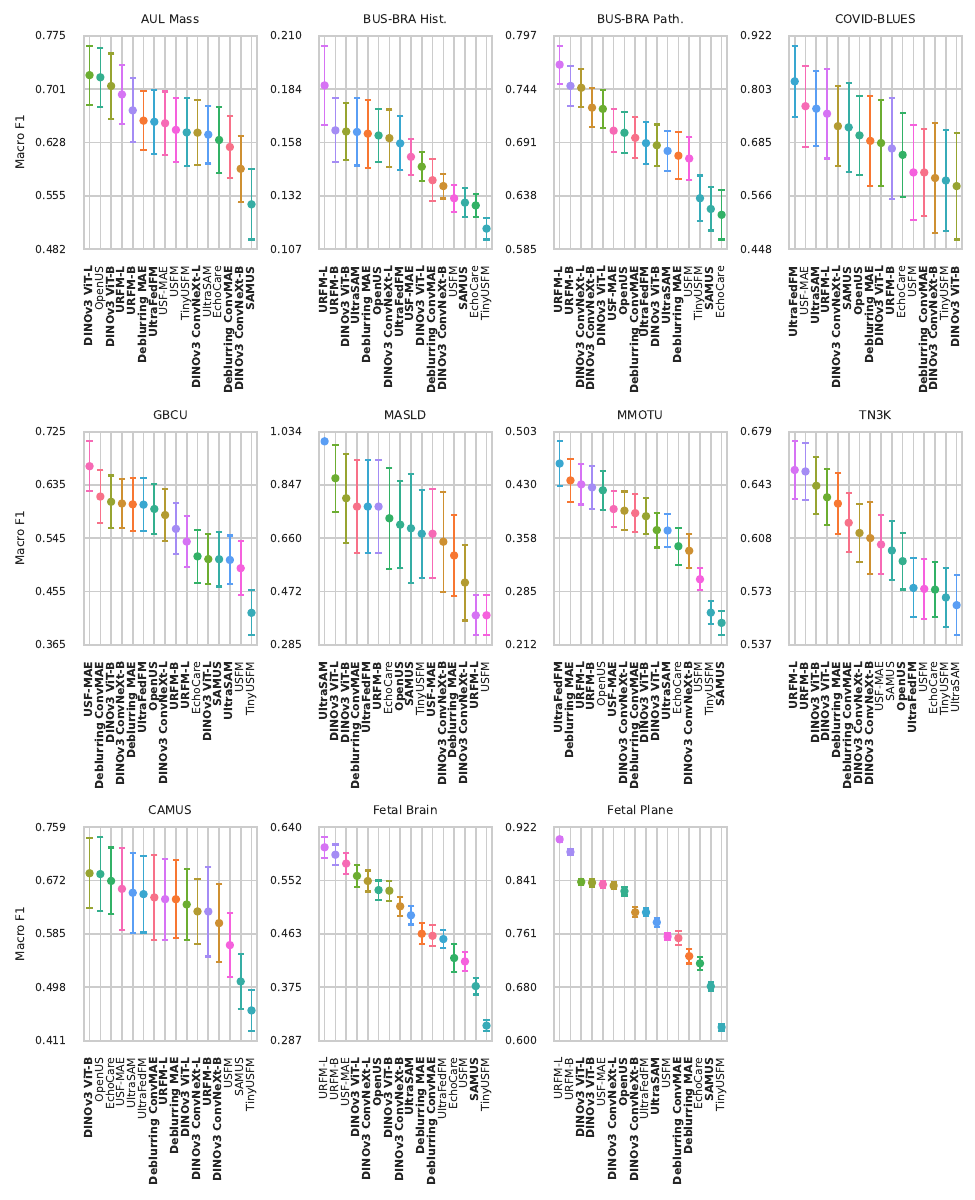}
    \caption{Head-only training results on each classification task. \textbf{Bold} indicates that the task dataset was not included in the model's pretraining data.}
    \label{fig:per_task_classification_head_only}
\end{figure*}

\begin{figure*}[!htbp]
    \centering
    \includegraphics[width=0.98\linewidth]{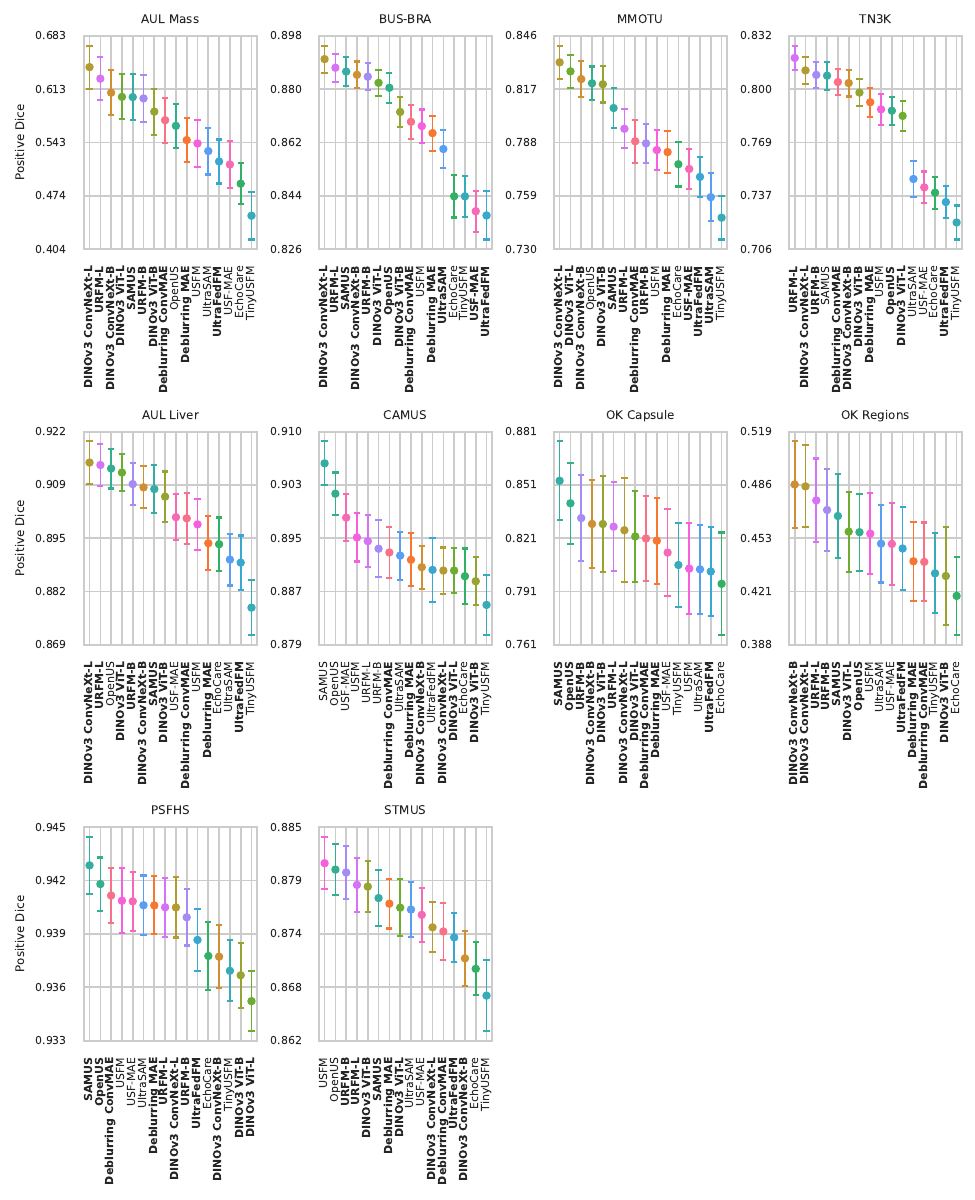}
    \caption{Full fine-tuning results on each segmentation task. \textbf{Bold} indicates that the task dataset was not included in the model's pretraining data.}
    \label{fig:per_task_segmentation_fft}
\end{figure*}

\begin{figure*}[!htbp]
    \centering
    \includegraphics[width=0.98\linewidth]{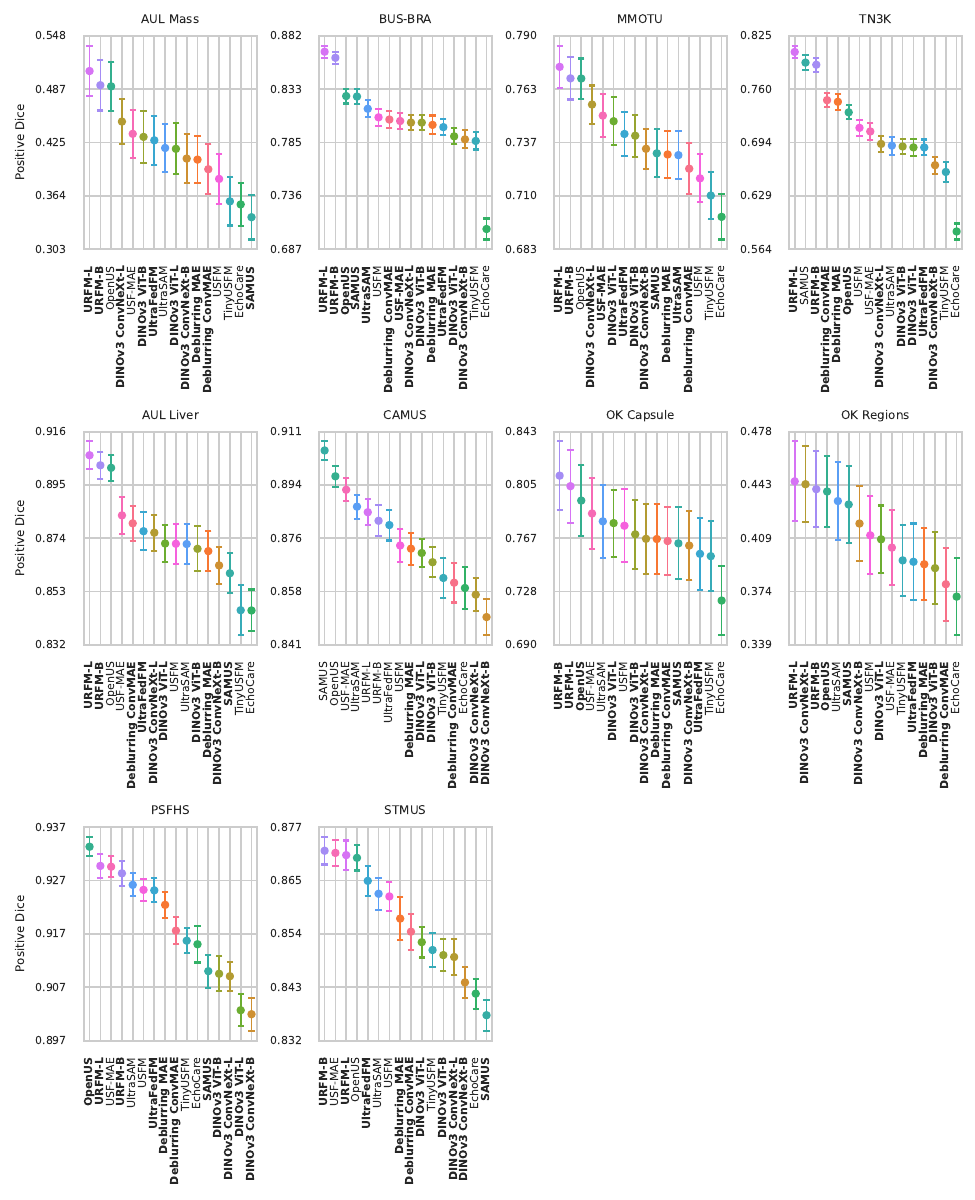}
    \caption{Head-only training results on each segmentation task. \textbf{Bold} indicates that the task dataset was not included in the model's pretraining data.}
    \label{fig:per_task_segmentation_head_only}
\end{figure*}

Figures~\ref{fig:per_task_classification_fft}, \ref{fig:per_task_classification_head_only}, \ref{fig:per_task_segmentation_fft}, and~\ref{fig:per_task_segmentation_head_only} give the per-task breakdown for both head-only and full fine-tuning training. Model names in bold are those whose pretraining corpus do not contain the task's dataset (see \appendixref{app:pretraining-overlap} for more information on overlap between pretraining corpora and UltraBench 2 evaluation datasets).

\section{Results Tables}

This appendix reports the numbers behind the figures in \appendixref{app:per-task-results}, giving the mean and standard error with the best result per task in bold. Tables~\ref{tab:classification_full_finetuning} and~\ref{tab:classification_head_only} report the macro F1 score on each classification task under full fine-tuning and head-only training respectively, and Tables~\ref{tab:segmentation_full_finetuning} and~\ref{tab:segmentation_head_only} report the positive Dice score on each segmentation task under the same two protocols. We include these for full transparency.

\begin{table*}[!bp]
\centering
\caption{The mean and standard error of the macro F1 score for each classification task after full fine-tuning. The best result for each task is shown in bold.}
\label{tab:classification_full_finetuning}
\fontsize{10}{12}\selectfont
\setlength{\tabcolsep}{3pt}
\sisetup{
  table-format = 1.3(3),
  uncertainty-mode = separate,
  retain-zero-uncertainty = true,
  mode = text,
  reset-text-series = false,
}
\resizebox{\textwidth}{!}{%
\begin{tabular}{lSSSSSSSSSSS}
\toprule
{} & \multicolumn{8}{c}{Pathological} & \multicolumn{3}{c}{Non-Pathological} \\
\cmidrule(lr){2-9} \cmidrule(lr){10-12}
{} & {AUL Mass} & {BUS-BRA Hist.} & {BUS-BRA Path.} & {COVID-BLUES} & {GBCU} & {MASLD} & {MMOTU} & {TN3K} & {CAMUS} & {Fetal Brain} & {Fetal Plane} \\
\midrule
Deblurring ConvMAE & 0.639(0.040) & 0.174(0.017) & 0.702(0.022) & 0.644(0.100) & \fontseries{b}\selectfont 0.741(0.040) & 0.756(0.149) & 0.452(0.026) & 0.692(0.020) & 0.680(0.055) & 0.534(0.020) & 0.835(0.006) \\
Deblurring MAE & 0.643(0.044) & 0.179(0.018) & 0.702(0.021) & 0.624(0.098) & 0.641(0.043) & 0.791(0.158) & 0.470(0.031) & 0.664(0.019) & 0.675(0.057) & 0.532(0.015) & 0.830(0.009) \\
DINOv3 ConvNeXt-B & 0.616(0.044) & 0.147(0.010) & 0.711(0.019) & 0.548(0.113) & 0.678(0.036) & 0.644(0.142) & 0.435(0.028) & 0.632(0.018) & 0.628(0.059) & 0.595(0.015) & 0.810(0.025) \\
DINOv3 ConvNeXt-L & 0.687(0.043) & 0.148(0.008) & 0.733(0.021) & 0.686(0.100) & 0.698(0.038) & 0.450(0.087) & 0.382(0.026) & 0.632(0.023) & 0.592(0.051) & 0.573(0.018) & 0.844(0.011) \\
DINOv3 ViT-B & 0.722(0.041) & 0.152(0.011) & 0.696(0.020) & 0.538(0.099) & 0.670(0.040) & 0.692(0.152) & 0.423(0.025) & 0.696(0.017) & 0.701(0.057) & 0.598(0.017) & 0.850(0.021) \\
DINOv3 ViT-L & \fontseries{b}\selectfont 0.724(0.038) & 0.156(0.007) & 0.735(0.023) & 0.585(0.110) & 0.631(0.040) & 0.804(0.138) & 0.366(0.027) & 0.664(0.021) & 0.639(0.055) & 0.623(0.017) & 0.868(0.007) \\
EchoCare & 0.614(0.051) & 0.130(0.006) & 0.668(0.021) & 0.627(0.098) & 0.647(0.050) & 0.765(0.165) & 0.352(0.027) & 0.586(0.022) & 0.698(0.055) & 0.535(0.034) & 0.781(0.009) \\
OpenUS & 0.707(0.039) & 0.153(0.013) & 0.710(0.021) & 0.766(0.094) & 0.696(0.040) & 0.708(0.151) & 0.458(0.029) & 0.654(0.019) & \fontseries{b}\selectfont 0.710(0.055) & 0.646(0.021) & 0.848(0.012) \\
SAMUS & 0.644(0.044) & 0.163(0.018) & 0.731(0.021) & 0.735(0.093) & 0.643(0.039) & 0.570(0.149) & 0.360(0.031) & 0.664(0.018) & 0.630(0.059) & 0.500(0.015) & 0.802(0.009) \\
TinyUSFM & 0.619(0.042) & 0.148(0.010) & 0.721(0.020) & 0.576(0.109) & 0.627(0.041) & 0.644(0.161) & 0.384(0.023) & 0.601(0.019) & 0.675(0.054) & 0.462(0.015) & 0.787(0.008) \\
UltraFedFM & 0.684(0.041) & 0.163(0.015) & 0.687(0.021) & \fontseries{b}\selectfont 0.804(0.084) & 0.643(0.037) & 0.831(0.154) & \fontseries{b}\selectfont 0.514(0.030) & 0.612(0.019) & 0.673(0.060) & 0.556(0.023) & 0.823(0.007) \\
UltraSam & 0.672(0.041) & 0.161(0.018) & 0.698(0.019) & 0.688(0.091) & 0.541(0.041) & \fontseries{b}\selectfont 0.916(0.096) & 0.407(0.025) & 0.585(0.020) & 0.639(0.062) & 0.581(0.018) & 0.817(0.009) \\
URFM-B & 0.662(0.043) & 0.188(0.014) & 0.774(0.019) & 0.663(0.112) & 0.679(0.039) & 0.771(0.163) & 0.490(0.028) & 0.737(0.018) & 0.631(0.065) & 0.684(0.017) & 0.915(0.003) \\
URFM-L & 0.719(0.038) & \fontseries{b}\selectfont 0.201(0.016) & \fontseries{b}\selectfont 0.794(0.018) & 0.695(0.106) & 0.658(0.038) & 0.870(0.118) & 0.486(0.029) & \fontseries{b}\selectfont 0.738(0.017) & 0.653(0.070) & \fontseries{b}\selectfont 0.688(0.016) & \fontseries{b}\selectfont 0.930(0.004) \\
USFM & 0.680(0.042) & 0.145(0.005) & 0.753(0.020) & 0.608(0.103) & 0.686(0.039) & 0.612(0.171) & 0.419(0.031) & 0.697(0.018) & 0.582(0.070) & 0.581(0.021) & 0.870(0.005) \\
USF-MAE & 0.652(0.045) & 0.167(0.011) & 0.706(0.021) & 0.740(0.088) & 0.656(0.040) & 0.771(0.163) & 0.464(0.027) & 0.644(0.019) & 0.680(0.057) & 0.663(0.016) & 0.863(0.004) \\
\bottomrule
\end{tabular}
}
\end{table*}

\begin{table*}[!bp]
\centering
\caption{The mean and standard error of the macro F1 score for each classification task after head-only training. The best result for each task is shown in bold.}
\label{tab:classification_head_only}
\fontsize{10}{12}\selectfont
\setlength{\tabcolsep}{3pt}
\sisetup{
  table-format = 1.3(3),
  uncertainty-mode = separate,
  retain-zero-uncertainty = true,
  mode = text,
  reset-text-series = false,
}
\resizebox{\textwidth}{!}{%
\begin{tabular}{lSSSSSSSSSSS}
\toprule
{} & \multicolumn{8}{c}{Pathological} & \multicolumn{3}{c}{Non-Pathological} \\
\cmidrule(lr){2-9} \cmidrule(lr){10-12}
{} & {AUL Mass} & {BUS-BRA Hist.} & {BUS-BRA Path.} & {COVID-BLUES} & {GBCU} & {MASLD} & {MMOTU} & {TN3K} & {CAMUS} & {Fetal Brain} & {Fetal Plane} \\
\midrule
Deblurring ConvMAE & 0.622(0.043) & 0.140(0.010) & 0.696(0.020) & 0.619(0.097) & 0.616(0.045) & 0.771(0.163) & 0.392(0.026) & 0.618(0.020) & 0.645(0.070) & 0.460(0.018) & 0.755(0.011) \\
Deblurring MAE & 0.658(0.040) & 0.163(0.016) & 0.678(0.023) & 0.689(0.100) & 0.603(0.045) & 0.599(0.143) & 0.436(0.029) & 0.631(0.020) & 0.642(0.064) & 0.464(0.017) & 0.727(0.011) \\
DINOv3 ConvNeXt-B & 0.592(0.046) & 0.137(0.006) & 0.726(0.019) & 0.606(0.122) & 0.604(0.041) & 0.647(0.175) & 0.341(0.023) & 0.608(0.024) & 0.603(0.063) & 0.509(0.016) & 0.794(0.007) \\
DINOv3 ConvNeXt-L & 0.642(0.044) & 0.160(0.014) & 0.746(0.019) & 0.721(0.089) & 0.585(0.044) & 0.504(0.133) & 0.395(0.026) & 0.612(0.019) & 0.622(0.053) & 0.551(0.017) & 0.834(0.006) \\
DINOv3 ViT-B & 0.706(0.045) & 0.164(0.014) & 0.689(0.021) & 0.588(0.119) & 0.607(0.044) & 0.800(0.156) & 0.388(0.025) & 0.643(0.019) & \fontseries{b}\selectfont 0.684(0.057) & 0.535(0.016) & 0.838(0.006) \\
DINOv3 ViT-L & \fontseries{b}\selectfont 0.721(0.040) & 0.147(0.007) & 0.725(0.019) & 0.684(0.096) & 0.510(0.042) & 0.870(0.118) & 0.369(0.024) & 0.635(0.019) & 0.633(0.058) & 0.559(0.018) & 0.839(0.005) \\
EchoCare & 0.632(0.045) & 0.128(0.005) & 0.620(0.025) & 0.658(0.093) & 0.515(0.045) & 0.730(0.178) & 0.347(0.025) & 0.574(0.018) & 0.672(0.055) & 0.423(0.023) & 0.717(0.010) \\
OpenUS & 0.718(0.041) & 0.162(0.013) & 0.701(0.020) & 0.701(0.088) & 0.595(0.042) & 0.707(0.153) & 0.423(0.026) & 0.593(0.019) & 0.683(0.061) & 0.536(0.016) & 0.825(0.007) \\
SAMUS & 0.544(0.048) & 0.129(0.007) & 0.625(0.021) & 0.719(0.099) & 0.510(0.046) & 0.694(0.192) & 0.242(0.017) & 0.600(0.020) & 0.508(0.045) & 0.377(0.014) & 0.682(0.006) \\
TinyUSFM & 0.642(0.047) & 0.117(0.005) & 0.636(0.023) & 0.601(0.112) & 0.419(0.038) & 0.675(0.155) & 0.256(0.015) & 0.569(0.020) & 0.460(0.033) & 0.312(0.009) & 0.620(0.006) \\
UltraFedFM & 0.657(0.044) & 0.158(0.013) & 0.691(0.021) & \fontseries{b}\selectfont 0.821(0.079) & 0.602(0.045) & 0.771(0.163) & \fontseries{b}\selectfont 0.459(0.030) & 0.575(0.020) & 0.650(0.063) & 0.455(0.015) & 0.794(0.006) \\
UltraSam & 0.639(0.040) & 0.163(0.016) & 0.683(0.020) & 0.760(0.083) & 0.509(0.041) & \fontseries{b}\selectfont 1.000(0.000) & 0.368(0.022) & 0.564(0.020) & 0.653(0.066) & 0.494(0.015) & 0.778(0.007) \\
URFM-B & 0.672(0.044) & 0.164(0.015) & 0.748(0.020) & 0.672(0.113) & 0.561(0.043) & 0.771(0.163) & 0.427(0.029) & 0.652(0.019) & 0.622(0.073) & 0.595(0.017) & 0.884(0.004) \\
URFM-L & 0.694(0.040) & \fontseries{b}\selectfont 0.186(0.019) & \fontseries{b}\selectfont 0.769(0.019) & 0.749(0.100) & 0.540(0.043) & 0.389(0.070) & 0.431(0.027) & \fontseries{b}\selectfont 0.653(0.019) & 0.642(0.066) & \fontseries{b}\selectfont 0.607(0.017) & \fontseries{b}\selectfont 0.904(0.003) \\
USFM & 0.646(0.044) & 0.131(0.006) & 0.675(0.021) & 0.619(0.106) & 0.495(0.046) & 0.389(0.070) & 0.302(0.015) & 0.574(0.020) & 0.567(0.052) & 0.418(0.016) & 0.757(0.006) \\
USF-MAE & 0.655(0.044) & 0.151(0.009) & 0.703(0.022) & 0.766(0.090) & \fontseries{b}\selectfont 0.667(0.042) & 0.675(0.156) & 0.397(0.024) & 0.604(0.019) & 0.659(0.067) & 0.580(0.017) & 0.836(0.005) \\
\bottomrule
\end{tabular}
}
\end{table*}

\begin{table*}[!bp]
\centering
\caption{The mean and standard error of the positive Dice score for each segmentation task after full fine-tuning. The best result for each task is shown in bold.}
\label{tab:segmentation_full_finetuning}
\fontsize{10}{12}\selectfont
\setlength{\tabcolsep}{3pt}
\sisetup{
  table-format = 1.3(3),
  uncertainty-mode = separate,
  retain-zero-uncertainty = true,
  mode = text,
  reset-text-series = false,
}
\resizebox{\textwidth}{!}{%
\begin{tabular}{lSSSSSSSSSS}
\toprule
{} & \multicolumn{4}{c}{Pathological} & \multicolumn{6}{c}{Anatomical} \\
\cmidrule(lr){2-5} \cmidrule(lr){6-11}
{} & {AUL Mass} & {BUS-BRA} & {MMOTU} & {TN3K} & {AUL Liver} & {CAMUS} & {OK Capsule} & {OK Regions} & {PSFHS} & {STMUS} \\
\midrule
Deblurring ConvMAE & 0.573(0.029) & 0.869(0.006) & 0.789(0.012) & 0.804(0.008) & 0.900(0.006) & 0.893(0.004) & 0.821(0.024) & 0.439(0.024) & 0.941(0.002) & 0.874(0.003) \\
Deblurring MAE & 0.547(0.029) & 0.865(0.006) & 0.783(0.011) & 0.792(0.008) & 0.894(0.007) & 0.892(0.004) & 0.820(0.024) & 0.439(0.024) & 0.941(0.002) & 0.877(0.003) \\
DINOv3 ConvNeXt-B & 0.609(0.029) & 0.885(0.004) & 0.823(0.010) & 0.804(0.008) & 0.908(0.005) & 0.891(0.003) & 0.829(0.025) & \fontseries{b}\selectfont 0.487(0.027) & 0.938(0.002) & 0.871(0.003) \\
DINOv3 ConvNeXt-L & \fontseries{b}\selectfont 0.642(0.028) & \fontseries{b}\selectfont 0.890(0.005) & \fontseries{b}\selectfont 0.832(0.009) & 0.811(0.008) & \fontseries{b}\selectfont 0.914(0.005) & 0.890(0.003) & 0.825(0.029) & 0.485(0.025) & 0.940(0.002) & 0.874(0.003) \\
DINOv3 ViT-B & 0.584(0.030) & 0.872(0.005) & 0.820(0.010) & 0.798(0.008) & 0.906(0.006) & 0.889(0.003) & 0.829(0.027) & 0.430(0.030) & 0.936(0.002) & 0.879(0.003) \\
DINOv3 ViT-L & 0.603(0.029) & 0.882(0.004) & 0.827(0.009) & 0.784(0.009) & 0.912(0.005) & 0.890(0.003) & 0.822(0.026) & 0.457(0.025) & 0.935(0.002) & 0.876(0.003) \\
EchoCare & 0.489(0.027) & 0.844(0.007) & 0.776(0.012) & 0.739(0.010) & 0.894(0.007) & 0.889(0.004) & 0.795(0.029) & 0.418(0.024) & 0.938(0.002) & 0.870(0.003) \\
OpenUS & 0.565(0.029) & 0.880(0.005) & 0.820(0.009) & 0.787(0.008) & 0.913(0.005) & 0.901(0.003) & 0.840(0.023) & 0.457(0.024) & 0.942(0.002) & 0.880(0.003) \\
SAMUS & 0.603(0.030) & 0.886(0.005) & 0.807(0.011) & 0.808(0.008) & 0.908(0.006) & \fontseries{b}\selectfont 0.906(0.003) & \fontseries{b}\selectfont 0.853(0.022) & 0.467(0.026) & \fontseries{b}\selectfont 0.943(0.002) & 0.877(0.003) \\
TinyUSFM & 0.448(0.031) & 0.844(0.007) & 0.747(0.012) & 0.721(0.010) & 0.878(0.007) & 0.885(0.004) & 0.806(0.024) & 0.432(0.025) & 0.937(0.002) & 0.867(0.004) \\
UltraFedFM & 0.519(0.029) & 0.837(0.008) & 0.769(0.011) & 0.734(0.010) & 0.889(0.007) & 0.890(0.005) & 0.802(0.025) & 0.447(0.025) & 0.938(0.002) & 0.873(0.003) \\
UltraSam & 0.532(0.031) & 0.860(0.006) & 0.758(0.013) & 0.747(0.010) & 0.890(0.006) & 0.892(0.003) & 0.804(0.025) & 0.450(0.024) & 0.941(0.002) & 0.876(0.003) \\
URFM-B & 0.601(0.031) & 0.884(0.005) & 0.788(0.010) & 0.809(0.008) & 0.909(0.005) & 0.893(0.004) & 0.832(0.024) & 0.471(0.025) & 0.940(0.002) & 0.880(0.003) \\
URFM-L & 0.627(0.028) & 0.887(0.005) & 0.796(0.011) & \fontseries{b}\selectfont 0.819(0.007) & 0.913(0.005) & 0.894(0.004) & 0.827(0.025) & 0.477(0.026) & 0.940(0.002) & 0.879(0.003) \\
USFM & 0.542(0.031) & 0.867(0.006) & 0.784(0.011) & 0.788(0.009) & 0.899(0.006) & 0.895(0.003) & 0.804(0.025) & 0.456(0.025) & 0.941(0.002) & \fontseries{b}\selectfont 0.881(0.003) \\
USF-MAE & 0.515(0.031) & 0.839(0.007) & 0.774(0.011) & 0.742(0.009) & 0.901(0.006) & 0.898(0.003) & 0.813(0.024) & 0.450(0.025) & 0.941(0.002) & 0.876(0.003) \\
\bottomrule
\end{tabular}
}
\end{table*}

\begin{table*}[!bp]
\centering
\caption{The mean and standard error of the positive Dice score for each segmentation task after head-only training. The best result for each task is shown in bold.}
\label{tab:segmentation_head_only}
\fontsize{10}{12}\selectfont
\setlength{\tabcolsep}{3pt}
\sisetup{
  table-format = 1.3(3),
  uncertainty-mode = separate,
  retain-zero-uncertainty = true,
  mode = text,
  reset-text-series = false,
}
\resizebox{\textwidth}{!}{%
\begin{tabular}{lSSSSSSSSSS}
\toprule
{} & \multicolumn{4}{c}{Pathological} & \multicolumn{6}{c}{Anatomical} \\
\cmidrule(lr){2-5} \cmidrule(lr){6-11}
{} & {AUL Mass} & {BUS-BRA} & {MMOTU} & {TN3K} & {AUL Liver} & {CAMUS} & {OK Capsule} & {OK Regions} & {PSFHS} & {STMUS} \\
\midrule
Deblurring ConvMAE & 0.395(0.029) & 0.806(0.008) & 0.724(0.013) & 0.746(0.009) & 0.880(0.007) & 0.861(0.007) & 0.764(0.024) & 0.378(0.024) & 0.917(0.003) & 0.855(0.004) \\
Deblurring MAE & 0.406(0.027) & 0.801(0.009) & 0.731(0.012) & 0.744(0.010) & 0.869(0.008) & 0.873(0.005) & 0.766(0.025) & 0.391(0.024) & 0.922(0.002) & 0.857(0.005) \\
DINOv3 ConvNeXt-B & 0.407(0.028) & 0.788(0.008) & 0.734(0.010) & 0.667(0.010) & 0.863(0.007) & 0.850(0.006) & 0.761(0.025) & 0.418(0.024) & 0.902(0.003) & 0.844(0.003) \\
DINOv3 ConvNeXt-L & 0.449(0.026) & 0.803(0.007) & 0.756(0.010) & 0.693(0.009) & 0.876(0.007) & 0.857(0.006) & 0.766(0.025) & 0.444(0.025) & 0.909(0.003) & 0.849(0.004) \\
DINOv3 ViT-B & 0.432(0.030) & 0.803(0.007) & 0.740(0.011) & 0.690(0.010) & 0.870(0.009) & 0.868(0.005) & 0.769(0.025) & 0.389(0.024) & 0.909(0.003) & 0.850(0.003) \\
DINOv3 ViT-L & 0.418(0.029) & 0.790(0.007) & 0.747(0.012) & 0.688(0.011) & 0.872(0.007) & 0.871(0.005) & 0.777(0.024) & 0.408(0.022) & 0.902(0.003) & 0.852(0.003) \\
EchoCare & 0.354(0.025) & 0.706(0.010) & 0.699(0.011) & 0.585(0.010) & 0.845(0.008) & 0.860(0.007) & 0.722(0.025) & 0.370(0.025) & 0.915(0.003) & 0.842(0.003) \\
OpenUS & 0.490(0.028) & 0.827(0.007) & 0.769(0.010) & 0.731(0.009) & 0.902(0.005) & 0.896(0.003) & 0.793(0.025) & 0.439(0.024) & \fontseries{b}\selectfont 0.933(0.002) & 0.870(0.003) \\
SAMUS & 0.339(0.026) & 0.827(0.007) & 0.731(0.012) & 0.792(0.009) & 0.860(0.008) & \fontseries{b}\selectfont 0.905(0.003) & 0.763(0.026) & 0.431(0.025) & 0.910(0.003) & 0.837(0.003) \\
TinyUSFM & 0.358(0.028) & 0.786(0.008) & 0.710(0.012) & 0.658(0.012) & 0.845(0.010) & 0.863(0.007) & 0.754(0.025) & 0.394(0.023) & 0.916(0.002) & 0.851(0.004) \\
UltraFedFM & 0.428(0.028) & 0.799(0.007) & 0.741(0.011) & 0.688(0.010) & 0.877(0.008) & 0.880(0.005) & 0.755(0.026) & 0.393(0.025) & 0.925(0.002) & 0.865(0.003) \\
UltraSam & 0.419(0.028) & 0.816(0.008) & 0.730(0.012) & 0.691(0.011) & 0.871(0.008) & 0.886(0.004) & 0.779(0.026) & 0.433(0.026) & 0.926(0.002) & 0.863(0.003) \\
URFM-B & 0.491(0.029) & 0.862(0.005) & 0.769(0.011) & 0.790(0.009) & 0.903(0.005) & 0.882(0.005) & \fontseries{b}\selectfont 0.811(0.025) & 0.441(0.025) & 0.928(0.002) & \fontseries{b}\selectfont 0.872(0.003) \\
URFM-L & \fontseries{b}\selectfont 0.508(0.029) & \fontseries{b}\selectfont 0.868(0.006) & \fontseries{b}\selectfont 0.775(0.011) & \fontseries{b}\selectfont 0.805(0.008) & \fontseries{b}\selectfont 0.906(0.005) & 0.885(0.004) & 0.804(0.026) & \fontseries{b}\selectfont 0.446(0.026) & 0.930(0.002) & 0.871(0.003) \\
USFM & 0.383(0.029) & 0.808(0.008) & 0.719(0.012) & 0.712(0.010) & 0.872(0.008) & 0.874(0.005) & 0.775(0.026) & 0.410(0.025) & 0.925(0.002) & 0.862(0.003) \\
USF-MAE & 0.435(0.028) & 0.804(0.008) & 0.750(0.011) & 0.708(0.010) & 0.883(0.007) & 0.892(0.004) & 0.784(0.025) & 0.402(0.025) & 0.930(0.002) & 0.871(0.003) \\
\bottomrule
\end{tabular}
}
\end{table*}

\section{Head Architectures}
\label{app:head-architectures}

This appendix specifies the task heads attached to each backbone. All heads are randomly initialized, and are the only trainable parameters under head-only training.

\paragraph{Progressive upsampler (PUP) decoder.} Patch tokens are first reshaped from $[B, N, d]$ to a 2D grid $[B, d, \sqrt{N}, \sqrt{N}]$, while spatial maps $[B, C, H, W]$ are passed directly. A $1{\times}1$ convolution projects to 256 channels, followed by four progressive transposed-convolution upsampling stages ($256{\to}128{\to}64{\to}32{\to}16$ channels, stride~2), each followed by a   residual double-convolution block with GELU activations. A final $1{\times}1$ classification convolution and, if necessary, bilinear interpolation to the target resolution complete the decoder.

\paragraph{SegFormer decoder.} Each scale's features are linearly projected to 256 dimensions, bilinearly upsampled to the finest scale, concatenated, and fused via a $1{\times}1$ convolution with batch normalization and ReLU, before a final bilinear upsample to the target resolution.

Per-model decoder assignments and trainable parameter counts are given in Tables~\ref{tab:cls_param_count} and~\ref{tab:seg_param_count}.

\begin{table*}[!bp]
\centering
\caption{The number of trainable parameters in the linear classification head added to each model backbone for a binary classification task.}
\label{tab:cls_param_count}
\small
\setlength{\tabcolsep}{3pt}
\sisetup{
  group-digits = true,
  group-separator = {,},
  table-format = 8.0,
}
\begin{tabular}{lSS}
\toprule
{Model} & {Feature Dimension} & {Trainable Parameters} \\
\midrule
Deblurring ConvMAE & 768 & 1538 \\
Deblurring MAE & 768 & 1538 \\
DINOv3 ConvNeXt-B & 1024 & 2050 \\
DINOv3 ConvNeXt-L & 1536 & 3074 \\
DINOv3 ViT-B & 768 & 1538 \\
DINOv3 ViT-L & 1024 & 2050 \\
EchoCare & 2048 & 4098 \\
OpenUS & 768 & 1538 \\
SAMUS & 1024 & 2050 \\
TinyUSFM & 192 & 386 \\
UltraFedFM & 768 & 1538 \\
UltraSam & 768 & 1538 \\
URFM-B & 768 & 1538 \\
URFM-L & 1024 & 2050 \\
USFM & 768 & 1538 \\
USF-MAE & 768 & 1538 \\
\bottomrule
\end{tabular}
\end{table*}

\begin{table*}[!bp]
\centering
\caption{The number of trainable parameters in each segmentation decoder head compatible with each model backbone's output type, for a binary segmentation task.}
\label{tab:seg_param_count}
\small
\setlength{\tabcolsep}{3pt}
\sisetup{
  group-digits = true,
  group-separator = {,},
  table-format = 8.0,
}
\begin{tabular}{lllS}
\toprule
{Model} & {Output Type} & {Head Architecture} & {Trainable Parameters} \\
\midrule
Deblurring ConvMAE & Patch Tokens & PUP & 764338 \\
Deblurring MAE & Patch Tokens & PUP & 764338 \\
DINOv3 ConvNeXt-B & Multi-Scale Feature Maps & SegFormer & 755714 \\
DINOv3 ConvNeXt-L & Multi-Scale Feature Maps& SegFormer & 1001474 \\
DINOv3 ViT-B & Patch Tokens & PUP & 764338 \\
DINOv3 ViT-L & Patch Tokens & PUP & 829874 \\
EchoCare & Multi-Scale Feature Maps & SegFormer & 1345794 \\
OpenUS & Multi-Scale Feature Maps & SegFormer & 632834 \\
SAMUS & Feature Maps & PUP & 633266 \\
TinyUSFM & Patch Tokens & PUP & 616882 \\
UltraFedFM & Patch Tokens & PUP & 764338 \\
UltraSam & Patch Tokens & PUP & 764338 \\
URFM-B & Patch Tokens & PUP & 764338 \\
URFM-L & Patch Tokens & PUP & 829874 \\
USFM & Patch Tokens & PUP & 764338 \\
USF-MAE & Patch Tokens & PUP & 764338 \\
\bottomrule
\end{tabular}
\end{table*}

\section{Pre-Training Dataset Overlap}
\label{app:pretraining-overlap}

Several of the evaluated models are pretrained on datasets included in UltraBench~2. This means that they are exposed to the images for certain tasks before we fine-tuned them (even if only for unsupervised training). Table~\ref{tab:pretrain-overlap} records, for every model whose pretraining datasets are available, which UltraBench~2 datasets appear among them. We use it to mark uncontaminated model--task pairs throughout the results, so that readers can distinguish between genuine transfer and prior exposure. Here, we omit three models because their release descriptions do not enumerate their pretraining data.

\begin{table*}[!bp]
\centering
\small
\setlength{\tabcolsep}{4pt}
\caption{Overlap between model pretraining corpora and the UltraBench
evaluation datasets. \yes{}~=~dataset included in pretraining;
\no{}~=~not included. EchoCare, TinyUSFM, and USFM are omitted: all
three are pretrained in part on publicly available data, but the
constituent datasets are not disclosed, so their overlap with
UltraBench cannot be determined. DINOv3 does not knowingly include any ultrasound images in the training data, but it is not auditable.}
\label{tab:pretrain-overlap}
\begin{tabular}{l cccccccccccc}
\toprule
Model
 & \rot{AUL} & \rot{BUS-BRA} & \rot{CAMUS} & \rot{COVID-BLUES}
 & \rot{Fetal Planes} & \rot{GBCU} & \rot{MASLD} & \rot{MMOTU}
 & \rot{Open Kidney} & \rot{PSFHS} & \rot{STMUS} & \rot{TN3K} \\
\midrule
DeblurringMIM & \no  & \no & \no  & \no  & \no  & \no & \no & \no  & \no  & \no  & \no  & \no  \\
% DINOv3        & \no  & \no & \no  & \no  & \no  & \no & \no & \no  & \no  & \no  & \no  & \no  \\
OpenUS        & \yes & \no & \yes & \no  & \no  & \no & \no & \yes & \no  & \no  & \yes & \no  \\
SAMUS         & \no  & \no & \yes & \no  & \no  & \no & \no & \no  & \no  & \no  & \no  & \yes \\
UltraFedFM    & \no  & \no & \yes & \no  & \yes & \no & \no & \no  & \no  & \no  & \no  & \no  \\
UltraSam      & \yes & \no & \yes & \no  & \no  & \no & \no & \no  & \yes & \yes & \yes & \yes \\
URFM          & \no  & \no & \yes & \no  & \yes & \no & \no & \no  & \no  & \no  & \no  & \no  \\
USF-MAE       & \yes & \no & \yes & \yes & \yes & \no & \no & \no  & \yes & \yes & \yes & \yes \\
\bottomrule
\end{tabular}
\end{table*}

\section{Progressive Upsampler Ablation for SegFormer Models}
\label{app:segmentation-decoder-ablations}

We used a SegFormer decoder for DINOv3 ConvNeXt-B/L, EchoCare, and OpenUS because they output intermediate feature maps at different scales (coarse-to-fine features) as opposed to single-resolution feature maps (e.g., ViTs). We believe that accounting for this potential architectural advantage is the fairest approach for the comparisons. However, we encourage future work to more deeply investigate combinations of encoders and decoder architecture to isolate their effects.

Tables \ref{tab:segmentation_decoder_ablation_full_finetuning} and \ref{tab:segmentation_decoder_ablation_head_only} show the performance using the progressive upsampler (PUP) for these four models. The results show that the advantages of the SegFormer decoder are largely limited to head-only training. For full fine-tuning, the gains and losses are small ($\leq 0.014$) and inconsistent. For all models, the differences are not large enough to move significantly up or down the rankings (at most 3 places, and still within the noise).

\begin{table*}[!bp]
\centering
\caption{The mean and standard error of the positive Dice score over the pathological and anatomical segmentation tasks for each model after full fine-tuning with the SegFormer and upsampler decoders. The better decoder for each model and category is shown in bold.}
\label{tab:segmentation_decoder_ablation_full_finetuning}
\small
\setlength{\tabcolsep}{3pt}
\sisetup{
  table-format = 1.3(3),
  uncertainty-mode = separate,
  retain-zero-uncertainty = true,
  mode = text,
  reset-text-series = false,
}
\begin{tabular}{lSSSS}
\toprule
{} & \multicolumn{2}{c}{Pathological} & \multicolumn{2}{c}{Anatomical} \\
\cmidrule(lr){2-3} \cmidrule(lr){4-5}
{} & {SegFormer} & {Upsampler} & {SegFormer} & {Upsampler} \\
\midrule
DINOv3 ConvNeXt-B & 0.780(0.008) & \fontseries{b}\selectfont 0.788(0.008) & 0.820(0.007) & \fontseries{b}\selectfont 0.822(0.007) \\
DINOv3 ConvNeXt-L & 0.794(0.008) & \fontseries{b}\selectfont 0.808(0.008) & \fontseries{b}\selectfont 0.822(0.007) & 0.821(0.008) \\
EchoCare & 0.712(0.008) & \fontseries{b}\selectfont 0.719(0.008) & 0.801(0.008) & \fontseries{b}\selectfont 0.805(0.007) \\
OpenUS & 0.763(0.008) & \fontseries{b}\selectfont 0.771(0.008) & \fontseries{b}\selectfont 0.822(0.007) & 0.821(0.006) \\
\bottomrule
\end{tabular}
\end{table*}

\begin{table*}[!bp]
\centering
\caption{The mean and standard error of the positive Dice score over the pathological and anatomical segmentation tasks for each model after head-only training with the SegFormer and upsampler decoders. The better decoder for each model and category is shown in bold.}
\label{tab:segmentation_decoder_ablation_head_only}
\small
\setlength{\tabcolsep}{3pt}
\sisetup{
  table-format = 1.3(3),
  uncertainty-mode = separate,
  retain-zero-uncertainty = true,
  mode = text,
  reset-text-series = false,
}
\begin{tabular}{lSSSS}
\toprule
{} & \multicolumn{2}{c}{Pathological} & \multicolumn{2}{c}{Anatomical} \\
\cmidrule(lr){2-3} \cmidrule(lr){4-5}
{} & {SegFormer} & {Upsampler} & {SegFormer} & {Upsampler} \\
\midrule
DINOv3 ConvNeXt-B & \fontseries{b}\selectfont 0.649(0.008) & 0.577(0.008) & \fontseries{b}\selectfont 0.773(0.007) & 0.729(0.007) \\
DINOv3 ConvNeXt-L & \fontseries{b}\selectfont 0.675(0.008) & 0.594(0.008) & \fontseries{b}\selectfont 0.784(0.008) & 0.741(0.007) \\
EchoCare & \fontseries{b}\selectfont 0.586(0.008) & 0.565(0.008) & \fontseries{b}\selectfont 0.759(0.008) & 0.741(0.007) \\
OpenUS & \fontseries{b}\selectfont 0.704(0.008) & 0.658(0.008) & \fontseries{b}\selectfont 0.806(0.007) & 0.778(0.007) \\
\bottomrule
\end{tabular}
\end{table*}

\section{Influence of Small Datasets on Ranks}
\label{app:rank-stability}

To measure the influence of the small datasets on the overall rankings, we performed a leave-one-out analysis on the tasks with fewer than 500 training examples: kidney region segmentation, kidney capsule segmentation, COVID-BLUES classification, and MASLD classification. As summarized in Table~\ref{tab:small_task_exclusion_summary}, removing any one of these tasks changes the overall scores only slightly and reshuffles ranks mainly among models that are already effectively tied. This stability is reflected in the high Kendall and Spearman rank correlation coefficients, where a value of 1 indicates identical rankings.

Table~\ref{tab:small_task_exclusion_anatomy_segmentation} shows how anatomical segmentation scores and ranks change when the kidney capsule or kidney region task is excluded. Without the kidney capsule task, DINOv3 ConvNeXt-L ($0.821 \pm 0.005$) narrowly overtakes SAMUS ($0.820 \pm 0.005$) as the top-ranked model, although the difference is negligible. Without the kidney region task, the scores of all models increase, and the top three models keep their positions.

Table~\ref{tab:small_task_exclusion_pathology_classification} shows the corresponding changes for pathological classification when the MASLD or COVID-BLUES task is excluded. URFM-L and URFM-B remain first and second in both cases. The largest shifts occur when MASLD is excluded. In this case, DINOv3 ConvNeXt-L rises from 13th to 8th, while UltraSam falls from 8th to 14th. As with segmentation, however, many of these mid-table changes fall within the error bars.

\begin{table*}[!bp]
\centering
\caption{The stability of the model rankings after full fine-tuning when each task of fewer than 500 training examples is excluded from its category, compared with the ranking over all of the category's tasks.}
\label{tab:small_task_exclusion_summary}
\fontsize{10}{12}\selectfont
\setlength{\tabcolsep}{3pt}
\resizebox{\textwidth}{!}{%
\begin{tabular}{llcccccc}
\toprule
Category & Excluded task & Kendall's $\tau$ & Spearman's $\rho$ & Models moved & Max $|\Delta|$ rank & Top-1 kept & Mean $|\Delta|$ score \\
\midrule
Anatomical (segmentation) & Open Kidney Capsule & 0.750 & 0.921 & 13 & 3 & No & 0.0018 \\
Anatomical (segmentation) & Open Kidney Regions & 0.850 & 0.941 & 11 & 4 & Yes & 0.0720 \\
Pathological (classification) & MASLD & 0.700 & 0.850 & 11 & 6 & Yes & 0.0219 \\
Pathological (classification) & COVID-BLUES & 0.783 & 0.926 & 13 & 3 & Yes & 0.0112 \\
\bottomrule
\end{tabular}
}
\end{table*}

\begin{table*}[!bp]
\centering
\caption{The mean and standard error of the anatomical (segmentation) category score (positive Dice) after full fine-tuning, with all tasks and with each task of fewer than 500 training examples excluded in turn. $\Delta$ is the change in rank from all tasks; a positive value means the model ranks higher without the task. The best score in each column is shown in bold.}
\label{tab:small_task_exclusion_anatomy_segmentation}
\small
\setlength{\tabcolsep}{3pt}
\sisetup{
  table-format = 1.3(3),
  uncertainty-mode = separate,
  retain-zero-uncertainty = true,
  mode = text,
  reset-text-series = false,
}
\begin{tabular}{lScSccScc}
\toprule
{} & \multicolumn{2}{c}{All tasks} & \multicolumn{3}{c}{w/o Open Kidney Capsule} & \multicolumn{3}{c}{w/o Open Kidney Regions} \\
\cmidrule(lr){2-3} \cmidrule(lr){4-6} \cmidrule(lr){7-9}
{} & {Score} & {Rank} & {Score} & {Rank} & {$\Delta$} & {Score} & {Rank} & {$\Delta$} \\
\midrule
SAMUS & \fontseries{b}\selectfont 0.826(0.007) & 1 & 0.820(0.005) & 3 & $-2$ & \fontseries{b}\selectfont 0.897(0.005) & 1 & 0 \\
OpenUS & 0.822(0.007) & 2 & 0.819(0.005) & 5 & $-3$ & 0.895(0.005) & 2 & 0 \\
URFM-L & 0.822(0.007) & 3 & 0.821(0.005) & 2 & $+1$ & 0.891(0.005) & 3 & 0 \\
DINOv3 ConvNeXt-L & 0.822(0.007) & 4 & \fontseries{b}\selectfont 0.821(0.005) & 1 & $+3$ & 0.889(0.006) & 5 & $-1$ \\
URFM-B & 0.821(0.007) & 5 & 0.819(0.005) & 6 & $-1$ & 0.891(0.005) & 4 & $+1$ \\
DINOv3 ConvNeXt-B & 0.820(0.007) & 6 & 0.819(0.006) & 4 & $+2$ & 0.887(0.005) & 7 & $-1$ \\
DINOv3 ViT-L & 0.815(0.007) & 7 & 0.814(0.005) & 8 & $-1$ & 0.887(0.005) & 8 & $-1$ \\
USF-MAE & 0.813(0.007) & 8 & 0.813(0.005) & 9 & $-1$ & 0.886(0.005) & 10 & $-2$ \\
USFM & 0.813(0.007) & 9 & 0.814(0.005) & 7 & $+2$ & 0.884(0.005) & 12 & $-3$ \\
DINOv3 ViT-B & 0.811(0.007) & 10 & 0.808(0.006) & 13 & $-3$ & 0.888(0.006) & 6 & $+4$ \\
Deblurring ConvMAE & 0.811(0.007) & 11 & 0.809(0.005) & 11 & 0 & 0.886(0.005) & 9 & $+2$ \\
Deblurring MAE & 0.810(0.007) & 12 & 0.808(0.005) & 12 & 0 & 0.885(0.005) & 11 & $+1$ \\
UltraSam & 0.809(0.007) & 13 & 0.810(0.005) & 10 & $+3$ & 0.881(0.005) & 13 & 0 \\
UltraFedFM & 0.807(0.007) & 14 & 0.808(0.005) & 14 & 0 & 0.879(0.005) & 14 & 0 \\
TinyUSFM & 0.801(0.007) & 15 & 0.800(0.005) & 16 & $-1$ & 0.875(0.005) & 16 & $-1$ \\
EchoCare & 0.801(0.008) & 16 & 0.802(0.005) & 15 & $+1$ & 0.877(0.006) & 15 & $+1$ \\
\bottomrule
\end{tabular}
\end{table*}

\begin{table*}[!bp]
\centering
\caption{The mean and standard error of the pathological (classification) category score (macro F1) after full fine-tuning, with all tasks and with each task of fewer than 500 training examples excluded in turn. $\Delta$ is the change in rank from all tasks; a positive value means the model ranks higher without the task. The best score in each column is shown in bold.}
\label{tab:small_task_exclusion_pathology_classification}
\small
\setlength{\tabcolsep}{3pt}
\sisetup{
  table-format = 1.3(3),
  uncertainty-mode = separate,
  retain-zero-uncertainty = true,
  mode = text,
  reset-text-series = false,
}
\begin{tabular}{lScSccScc}
\toprule
{} & \multicolumn{2}{c}{All tasks} & \multicolumn{3}{c}{w/o MASLD} & \multicolumn{3}{c}{w/o COVID-BLUES} \\
\cmidrule(lr){2-3} \cmidrule(lr){4-6} \cmidrule(lr){7-9}
{} & {Score} & {Rank} & {Score} & {Rank} & {$\Delta$} & {Score} & {Rank} & {$\Delta$} \\
\midrule
URFM-L & \fontseries{b}\selectfont 0.645(0.021) & 1 & \fontseries{b}\selectfont 0.613(0.018) & 1 & 0 & \fontseries{b}\selectfont 0.638(0.019) & 1 & 0 \\
URFM-B & 0.620(0.026) & 2 & 0.599(0.019) & 2 & 0 & 0.614(0.025) & 2 & 0 \\
UltraFedFM & 0.617(0.023) & 3 & 0.587(0.016) & 4 & $-1$ & 0.590(0.024) & 4 & $-1$ \\
OpenUS & 0.606(0.023) & 4 & 0.592(0.016) & 3 & $+1$ & 0.584(0.024) & 6 & $-2$ \\
Deblurring ConvMAE & 0.600(0.024) & 5 & 0.578(0.018) & 5 & 0 & 0.594(0.024) & 3 & $+2$ \\
USF-MAE & 0.600(0.025) & 6 & 0.575(0.016) & 6 & 0 & 0.580(0.025) & 8 & $-2$ \\
Deblurring MAE & 0.589(0.025) & 7 & 0.561(0.018) & 10 & $-3$ & 0.584(0.025) & 5 & $+2$ \\
UltraSam & 0.584(0.019) & 8 & 0.536(0.018) & 14 & $-6$ & 0.569(0.017) & 11 & $-3$ \\
DINOv3 ViT-L & 0.583(0.025) & 9 & 0.551(0.021) & 12 & $-3$ & 0.583(0.023) & 7 & $+2$ \\
USFM & 0.575(0.026) & 10 & 0.570(0.018) & 7 & $+3$ & 0.570(0.026) & 10 & 0 \\
DINOv3 ViT-B & 0.574(0.025) & 11 & 0.557(0.017) & 11 & 0 & 0.579(0.023) & 9 & $+2$ \\
SAMUS & 0.564(0.024) & 12 & 0.563(0.017) & 9 & $+3$ & 0.539(0.023) & 13 & $-1$ \\
DINOv3 ConvNeXt-L & 0.552(0.020) & 13 & 0.567(0.018) & 8 & $+5$ & 0.533(0.015) & 16 & $-3$ \\
DINOv3 ConvNeXt-B & 0.551(0.024) & 14 & 0.538(0.020) & 13 & $+1$ & 0.552(0.022) & 12 & $+2$ \\
EchoCare & 0.549(0.025) & 15 & 0.518(0.016) & 16 & $-1$ & 0.537(0.025) & 14 & $+1$ \\
TinyUSFM & 0.540(0.025) & 16 & 0.525(0.018) & 15 & $+1$ & 0.535(0.024) & 15 & $+1$ \\
\bottomrule
\end{tabular}
\end{table*}

\section{Pathological and Anatomical Segmentation Predictions}
\label{app:segmentation-examples}

\begin{figure*}[!bp]
    \centering
    \includegraphics[width=\linewidth]{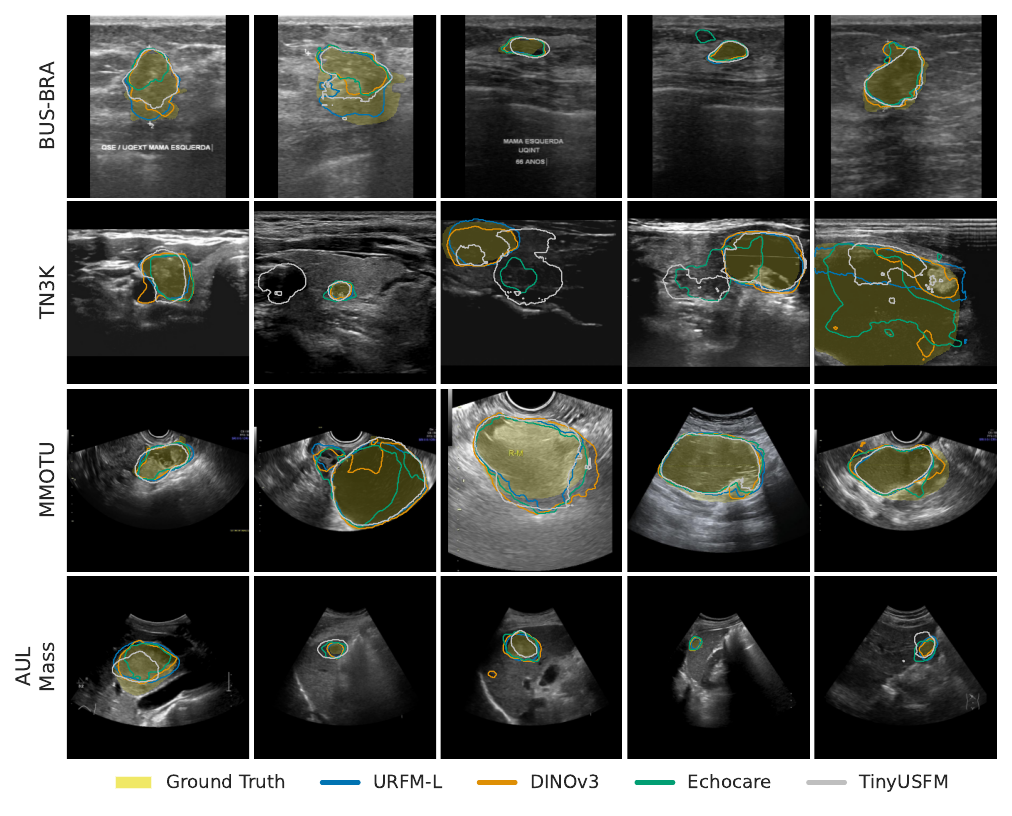}
    \caption{Examples of predicted segmentation boundaries for the best (URFM-L and DINOv3 ConvNeXt-L) and worst (EchoCare and TinyUSFM) pathological segmentation models.}
    \label{fig:pathological_segmentation_examples}
\end{figure*}

\begin{figure*}[!bp]
    \centering
    \includegraphics[width=\linewidth]{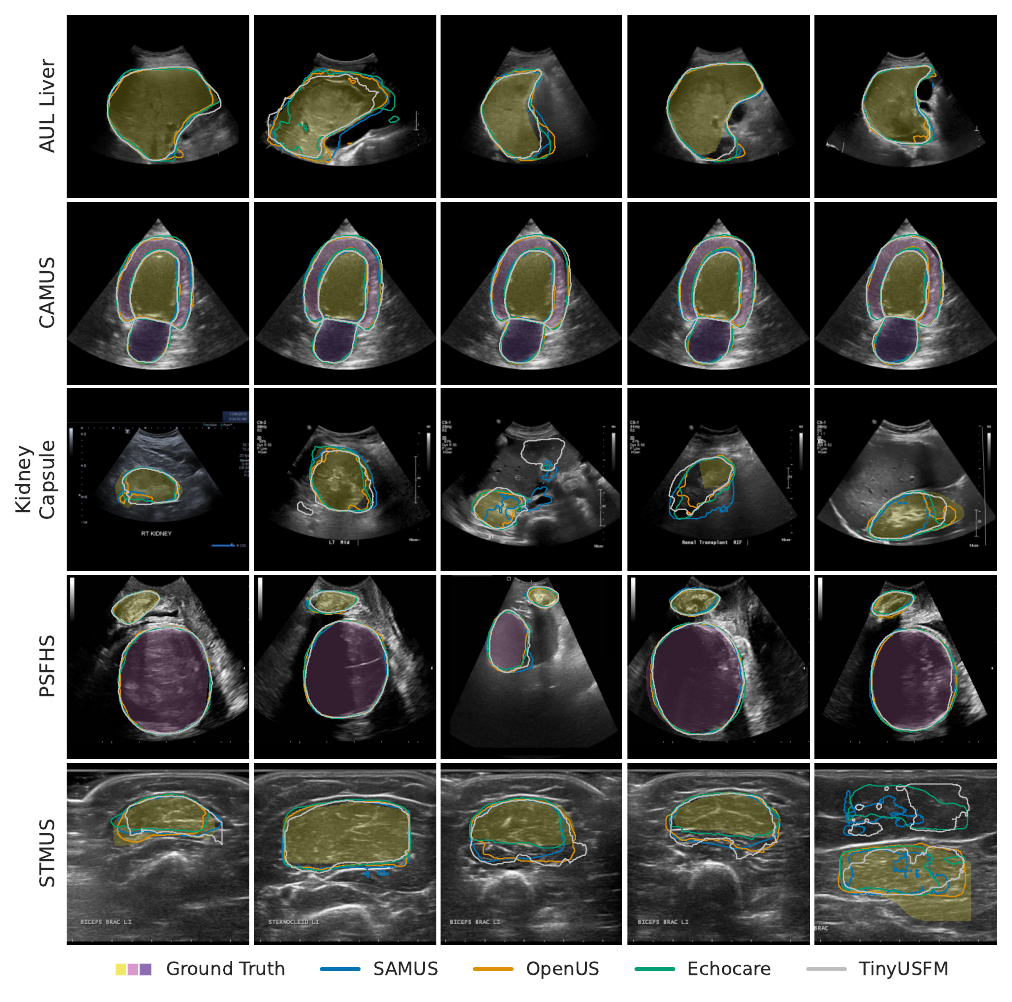}
    \caption{Examples of predicted segmentation boundaries for the best (SAMUS and OpenUS) and worst (EchoCare and TinyUSFM) anatomical segmentation models.}
    \label{fig:anatomical_segmentation_examples}
\end{figure*}

\begin{figure*}[!bp]
    \centering
    \includegraphics[width=\linewidth]{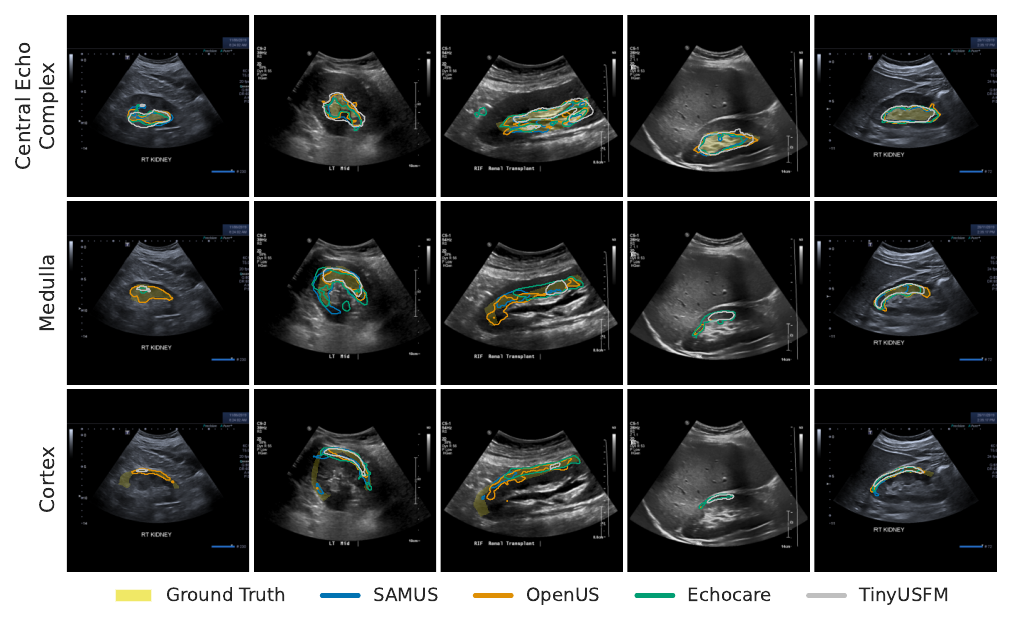}
    \caption{Examples of predicted segmentation boundaries for the best (SAMUS and OpenUS) and worst (EchoCare and TinyUSFM) anatomical segmentation models for each class on the Open Kidney regions task.}
    \label{fig:kidney_region_segmentation_examples}
\end{figure*}

Figures~\ref{fig:pathological_segmentation_examples}, \ref{fig:anatomical_segmentation_examples}, and~\ref{fig:kidney_region_segmentation_examples} compare the predicted segmentation boundaries of the best- and worst-performing pathological and anatomical segmentation models against the ground truth, using five test-set examples per task.

\end{document}